%% file: acl_submission.tex
\pdfoutput=1

\documentclass[11pt]{article}

\usepackage[final]{acl}

\usepackage{times}
\usepackage{latexsym}

\usepackage{times}
\usepackage{latexsym}
\usepackage[nolist]{acronym}
\usepackage{amssymb}
\usepackage{amsmath} 
\usepackage[capitalize]{cleveref}

\usepackage[utf8]{inputenc}

\usepackage{graphicx}
\usepackage{amsmath}
\usepackage[inkscapelatex=false]{svg}
\usepackage{mathtools}
\usepackage{amsfonts}
\usepackage{hhline}
\usepackage{makecell}
\usepackage{footnote}
\usepackage{booktabs}
\usepackage{array}
\usepackage{multirow}
\usepackage{subcaption}
\usepackage{afterpage}
\usepackage{url}
\usepackage{physics}
\makesavenoteenv{tabular}
\usepackage{subcaption}
\usepackage{hyperref}
\usepackage{url}
\usepackage{xspace}
\usepackage{enumitem}
\usepackage{xcolor}
\usepackage{colortbl}

\usepackage[T1]{fontenc}

\usepackage[utf8]{inputenc}

\usepackage{microtype}

\usepackage{inconsolata}

\usepackage{graphicx}

\usepackage{amsthm}

\newcommand{\saeki}{SAE$_g^G$\xspace}
\newcommand{\saei}[1]{SAE$_{#1}$\xspace}

\theoremstyle{plain}

\theoremstyle{definition}

\theoremstyle{remark}

\title{Group-SAE: Efficient Training of Sparse Autoencoders for Large Language Models via Layer Groups}

\author{
 \textbf{Davide Ghilardi\textsuperscript{1 *}},
 \textbf{Federico Belotti\textsuperscript{1 *}},
 \textbf{Marco Molinari\textsuperscript{2 *}},
 \textbf{Tao Ma\textsuperscript{2}},
 \textbf{Matteo Palmonari\textsuperscript{1}}
\\
\\
 \textsuperscript{1}University of Milan-Bicocca,
 \textsuperscript{2}London School of Economics \\
 \textsuperscript{* Equal contribution}
\\
 \small{
   \textbf{Correspondence:} \href{mailto:davide.ghilardi@unimib.it}
   {davide.ghilardi@unimib.it}
 }
}

\input{acronym}
\begin{document}
\maketitle
\begin{abstract}
\acp{sae} have recently been employed as a promising unsupervised approach for understanding the representations of layers of Large Language Models (LLMs). 
However, with the growth in model size and complexity, training \acp{sae} is computationally intensive, as typically one SAE is trained for each model layer.
To address such limitation, we propose \textit{Group-SAE}, a novel strategy to train SAEs. Our method considers the similarity of the residual stream representations between contiguous layers to group similar layers and train a single SAE per group. To balance the trade-off between efficiency and performance, we further introduce \textit{AMAD} (Average Maximum Angular Distance), an empirical metric that guides the selection of an optimal number of groups based on representational similarity across layers. Experiments on models from the Pythia family show that our approach significantly accelerates training with minimal impact on reconstruction quality and comparable downstream task performance and interpretability over baseline SAEs trained layer by layer.
This method provides an efficient and scalable strategy for training \acp{sae} in modern LLMs.
\end{abstract}

\section{Introduction}
\label{sec:intro}
\begin{figure}[ht!]
    \centering
    \includegraphics[width=1\linewidth]{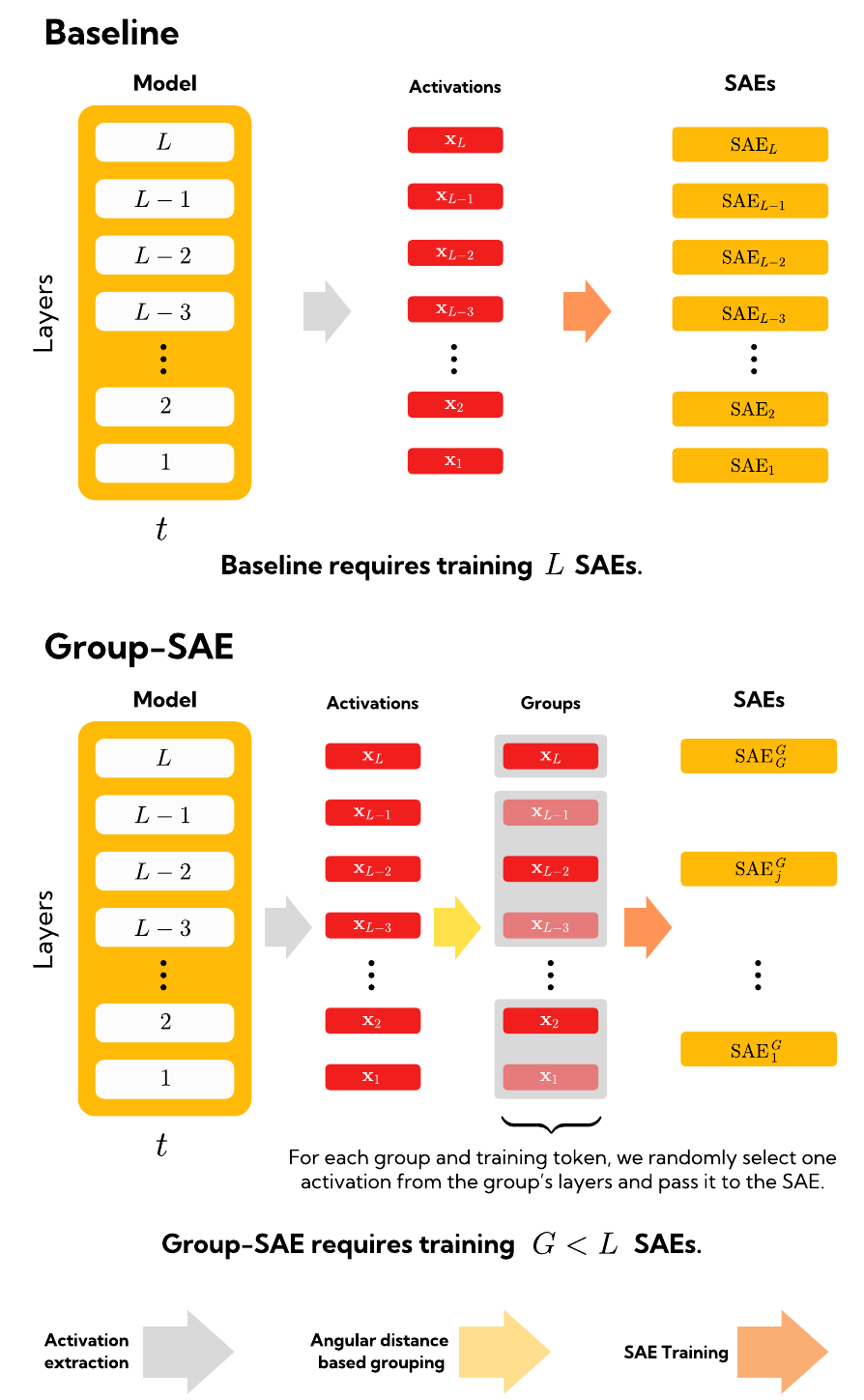}
    \caption{The illustration of our method. While standard training of \acp{sae} requires training one per layer, our method first groups layers by angular similarity and then trains a single \ac{sae} for each group.}
    \label{fig:method}
\end{figure}
Sparse Autoencoders (\acp{sae}) {~\cite{makhzani2014ksparse} have recently emerged   ~\cite{huben2024sparse, bricken2023monosemanticity}} as a promising technique {to tackle} the polysemanticity of neurons in the activations of \acp{llm} ~\cite{olah2020zoom}. \acp{sae} decompose models' activations into a sparse combination of human-interpretable directions, also called \emph{features}. Despite the strengths in interpretability, \acp{sae} {face} challenges that hinder their large-scale adoption \cite{sharkey2025openproblems}. One of them is the high training and evaluation costs, which increase as model sizes and parameter counts grow. Notably, a separate \ac{sae} is typically trained for each component {(e.g., the output of the attention, the MLP, or a full transformer block)} at every layer of an \ac{llm}, with a number of features that is a multiple of the dimensionality of the activation space of the model. 
For instance, a single \ac{sae} trained on the activations of a layer of Llama-3.1 8B \citep{grattafiori2024llama3herdmodels}, with an expansion factor of 32, involves approximately ${4096^2 \times 32 \times 2 \approx 1.073}$ billion parameters. 
Such high computational demand increases training time and requires substantial hardware resources and energy consumption, making the approach increasingly impractical as models scale. Moreover, to make \acp{sae} useful \textcolor{black}{for interpretability}, all their features have to be manually annotated. Even when using auto-interpretability techniques, this process can become \textcolor{black}{unsustainable} \cite{paulo2024auto-interp}.

Facing such challenges, in this work we introduce \textbf{Group-SAE}, \textcolor{black}{depicted in Figure \ref{fig:method}}, a method to reduce the computational overhead of training, evaluating, and interpreting \acp{sae}. Our method leverages the similarity of the representations shared by close layers to reduce the total number of trained \acp{sae} and uses a single SAE to reconstruct activations from different layers. The proposed technique follows primary observations that nearby neural network layers tend to learn similar levels of representations \cite{szegedy2014intriguingpropertiesneuralnetworks, zeiler2014visualizing, jawahar-etal-2019-bert}. Shallow layers typically focus on capturing low-level features, while deeper layers are believed to learn high-level abstractions. In addition, \citet{gromov2024unreasonable} empirically shows that adjacent layers in \ac{llm}s could encode similar information.

Additionally, we introduce \textbf{AMAD} (Average Maximum Angular Distance), a novel empirical metric for selecting the optimal number of groups to partition a model's layers--an important choice that balances SAEs quality and computational efficiency: more groups tend to improve reconstruction performance but reduce computational savings, while fewer groups offer greater efficiency at the cost of decreasing the quality of the reconstruction.

\textcolor{black}{After thoroughly evaluating the reconstruction, downstream, and interpretability results of our methods on three models of varying sizes from the Pythia family \cite{biderman2023pythia}--{Pythia-160M}, {Pythia-410M}, and {Pythia-1B}--we demonstrate that our method has several advantages over baselines.} In particular, \textcolor{black}{Group-SAE with AMAD} finds an optimal tradeoff between training costs and quality of the SAE.
It significantly reduces the number of trained \acp{sae}, reducing training costs up to 50\%. Moreover, such a novel approach only incurs a slight decrease in reconstruction quality and achieves comparable downstream performance. \textcolor{black}{Additionally, Group-SAEs outperform standard \acp{sae} matching the same computational cost.} Finally, from an interpretability point of view, Group-\acp{sae} offers the same, or even slightly better, level of interpretability when compared with their baseline counterparts. Our \textbf{contributions} can be summarized as follows:
\begin{itemize}[labelindent=0em, leftmargin=1.3em]
    \item We propose a novel method named \textbf{Group-SAE}, which partitions the layers of a model into groups and trains a single \ac{sae} for each group, thus significantly reducing the total number of \acp{sae} to train.
    \item We introduce \textbf{AMAD} (Average Maximum Angular Distance), a new empirical metric for selecting the optimal number of groups, enabling an effective trade-off between computational efficiency and performance.
\end{itemize}

All the SAEs trained and used in our experiments, the code to train Group-SAE, and the code to replicate the experiments are all released as open source at \href{https://github.com/ghidav/group-sae}{\url{https://github.com/ghidav/group-sae}}

\section{Background and Related Work}
\label{sec:sota}

\subsection{Sparse Autoencoders}

\acp{sae}~\cite{bricken2023monosemanticity} are a promising interpretability technique that decomposes dense \ac{llm} activations into a sparse combination of human-interpretable features.
\acp{sae} are based on two key intuitions. The first is the \ac{lrh}, which, supported by substantial empirical evidence~\cite{mikolov-etal-2013-linguistic, nanda-etal-2023-emergent, park2023thelinearreprhypandgeometry}, posits that \acp{nn} exhibit interpretable linear directions in their activation space. The second is the \ac{sh}, which assumes that observed \acp{nn} are dense compressions of a larger sparse model where each neuron corresponds to a specific feature~\cite{elhage2022superposition}.

Within this framework, \acp{sae} {disentangle} the effects of superposition, enabling the learning of interpretable linear directions in the model's activations. Formally, given an activation $\mathbf{x} \in \mathbb{R}^n$, a \ac{sae} reconstructs it through two steps. First, it encodes the activation into the feature space as:
\begin{equation}
    \mathbf{f}(\mathbf{x}) =
    \sigma\,\big(\mathbf{b}_{e} + \mathbf{W}_{e}\,(\mathbf{x} - \mathbf b_d)\big)
    \label{eqn:encode}
\end{equation}
where $\mathbf{f}(\mathbf{x})$ represents feature activations, ${\mathbf{b}_{e} \in \mathbb{R}^m},{\mathbf{b}_{d} \in \mathbb{R}^n}$ are bias terms, $\mathbf{W}_{e} \in \mathbb{R}^{m\times n}$ is the encoder matrix, and $\sigma$ is an activation function. Typically, $m=c \cdot n$, with the expansion factor ${c \in \{2^k \mid k \in \mathbb{N}_+\}}$. ${\sigma=\text{ReLU}}$ was initially proposed~\cite{bricken2023monosemanticity}, with its limitations that led to the development of two notable alternatives: TopK~\cite{gao2024topk} and JumpReLU~\cite{rajamanoharan2024jumprelu}.

The feature vector is then projected back into the model's activation space:
\begin{equation}
    \widehat{\mathbf x} = \mathbf{b}_{d} + \mathbf{W}_{d} \,\mathbf{f}(\mathbf{x}) 
    \label{eqn:decode}
\end{equation}
where $\mathbf{W}_{d} \in \mathbb{R}^{n\times m}$ is the decoder matrix, with each column corresponding to a learned feature vector. 

\acp{sae} are trained to minimize the MSE between original activations and \ac{sae} reconstruction. To enforce feature sparsity, an additional penalty is usually included in the loss function, either as the $L_1$ norm~\cite{bricken2023monosemanticity} or the $L_0$ norm~\cite{rajamanoharan2024jumprelu} of $\mathbf{f}(\mathbf{x})$, scaled by a positive factor $\lambda$, termed the \emph{sparsity coefficient}. Formally, the loss function can be written as:
\begin{equation}
    \mathcal L(\mathbf x)=\|\mathbf{x} - \widehat{\mathbf x}\|_2^2+\lambda\,\|\mathbf{f}(\mathbf x)\|_s
    \label{eqn:loss}
\end{equation}
with $s\in\{0,1\}$. {On the other hand}, when using TopK~\cite{gao2024topk}, no additional loss components are needed, as the activation function inherently enforces sparsity.

\subsection{Shared \acp{sae}}
\label{sec:shared-sae}
While \acp{sae} were originally designed to reconstruct activations from a single model component (e.g., the output of a specific layer, MLP, or Attention), subsequent approaches have explored their application to activations from multiple layers. For instance, \citet{yun2023dictionarylearning} and \citet{lawson2024mlsae} employed a single \ac{sae} to reconstruct activations from all residual stream layers of a model, aiming to analyze how features evolve across layers. More recently, \citet{lindsey2024crosscoders} extended this concept by introducing \emph{Crosscoder}, a modified \ac{sae} architecture that creates a unified representation of computations across multiple layers.

These methods are driven by empirical evidence suggesting that information in \acp{llm} is often shared and rather redundant across nearby layers~\cite{phang2021similarity, gromov2024unreasonable}. In this work, we leverage this principle to explore the optimal balance between performance and computational efficiency when applying \acp{sae} to multiple layers.

\subsection{Improving SAE efficiency}
\label{sec:sae-efficiency}

As highlighted by \citet{sharkey2025openproblems}, one of the major challenges of \acp{sae} is their high training and evaluation costs. As previously mentioned, \acp{sae} scale alongside model size, making them impractical for low-resource settings. Furthermore, interpreting the meaning of \ac{sae} features presents an additional challenge. Even with automated techniques, interpretation costs can reach thousands of dollars~\cite{paulo2024auto-interp}.

To mitigate training costs, \citet{gao2024topk} investigated the scaling laws of \acp{sae} to determine the optimal balance between model size and sparsity. Recent work has also explored transfer learning as a means to enhance \ac{sae} training efficiency. For instance, \citet{Kissane2024SAEs} and \citet{lieberum-etal-2024-gemma} demonstrated that \ac{sae} weights can be transferred between base and instruction-tuned versions of Gemma-1~\cite{gemmateam2024gemmaopenmodelsbased} and Gemma-2~\cite{gemmateam2024gemma2improvingopen}, respectively. Additionally, \citet{ghilardi-etal-2024-accelerating} showed that transferability also occurs within different layers of a single model, both in forward and backward directions.

\section{\textcolor{black}{Group-SAE}}

\textcolor{black}{In our approach, a \textbf{Group-SAE} is defined as a sparse autoencoder that is trained to reconstruct the activations from multiple layers that have been grouped together, rather than training an individual SAE for each layer. This grouping leverages the observation that nearby layers tend to exhibit similar activation patterns \cite{gromov2024unreasonable}. A detailed analysis of this phenomenon can be found in Appendix~\ref{app:ang-dist} (Figures~\ref{fig:pythia_160m_angular_distances}, \ref{fig:pythia_410m_angular_distances}, and \ref{fig:pythia_1b_angular_distances}).}

\subsection{\textcolor{black}{Clustering layers into groups}}
\label{sec:groups}

For a model with \(L\) layers, there are theoretically \(G! \cdot S(L, G)\) ways to partition the layers into \(G\) groups---where \(S(L, G)\) denotes the Stirling number of the second kind. Because this number grows rapidly with model depth, we instead employ an agglomerative clustering strategy based on angular distances between layers to efficiently determine a suitable grouping. Specifically, \textcolor{black}{following the formulation in \cite{gromov2024unreasonable}}, we compute the mean angular distance between the residual activations of each layer using \textcolor{black}{a subset of the training set used to train the SAEs} (see Appendix~\ref{app:ang-dist} for detailed measurements)~\footnote{We precisely use 10M tokens from the training set used to train the SAEs, which amount to 1\% of the total training tokens.}. We then apply a bottom-up hierarchical clustering method with complete linkage~\cite{nielsen2016hc}. At each step, the two groups with the smallest inter-group distance\footnote{In complete linkage, the inter-group distance is defined as
${D(X,Y)=\max_{\mathbf{x} \in X,\, \mathbf{y} \in Y} d_{\text{angular}}(\mathbf{x},\mathbf{y})}$
for groups \(X\) and \(Y\)}are merged. This merging continues until exactly \(G\) groups remain, ensuring that within each group the maximum angular distance is minimized.

\textcolor{black}{In addition to being motivated by recent work~\citep{gromov2024unreasonable, li2025mixln}, we adopt angular distance as our similarity metric because it captures the directional component of activations, which is key to their sparse representation. As in our setting, feature directions are typically normalized to unit norm, so SAE feature activations are proportional to the cosine of that angle. Consequently, activations that are close in angular distance tend to activate similar features and can be effectively reconstructed by the same SAE. Empirical evidence supports this: reconstruction quality degrades when SAEs are applied to activations from more distant layers~\citep{ghilardi-etal-2024-accelerating}, and this degradation correlates with larger angular distances.}

\subsection{\textcolor{black}{Selecting the number groups $G$}}
The choice of $G$, the number of groups of layers, is an important choice to make in our method as it influences both computational savings and the quality of the SAE. To guide this choice, we propose an empirical score called the \textbf{Average Maximum Angular Distance} (AMAD), defined as
\begin{equation}
\label{eqn:amad}
\text{AMAD}(G) = \frac{1}{G} \sum_{g=1}^{G} D_g,
\end{equation}
where $D_g$ is the maximum angular distance between any pair of activations within group $g$. AMAD thus quantifies, on average, the worst-case dissimilarity within groups. Intuitively, when $G$ is small, each group aggregates more distant layers, which increases AMAD; conversely, when $G=L$ (one group per layer), AMAD becomes zero but no computational savings are achieved. The goal is therefore to select the smallest $G$ such that the groups remain sufficiently homogeneous, i.e., AMAD$(G)$ stays below a target threshold. Formally, we select
\[
\widehat{G} = \min \{ G \mid \text{AMAD}(G) < \theta \},
\]
where $\theta$ is a distance threshold. Based on our empirical analysis (Section~\ref{sec:experiments}), we set $\theta = 0.2$, which provides a robust trade-off across model sizes. This criterion ensures maximal grouping (hence computational savings) while avoiding the sharp increase in reconstruction error that arises when groups mix overly distant layers.

\subsection{Computational Savings}
\label{sec:complexity}

The computational cost, in FLOPs, of training a \ac{sae} can be divided into two main components:
\begin{itemize}
    \item \emph{Activation caching ($A$)}: \textcolor{black}{FLOPs} required to generate the model's activations, which are used for training the \ac{sae}.
    \item \emph{SAE training ($T$)}: \textcolor{black}{FLOPs} involved in optimizing a \textcolor{black}{single} \ac{sae} using the cached activations.
\end{itemize}
Thus, the total cost of training \acp{sae} across all residual stream layers of a model is given by ${A + L T}$. Since both baseline and Group-\acp{sae} share the same architecture and undergo the same training process for a single \ac{sae}, the total cost of training all Group-\acp{sae} is ${A + G T}$~\footnote{We do not account for the cost of computing angular distance when selecting groups, as we rely on activations already sampled for training, making the additional computational overhead negligible.}.

The resulting compute savings, $\Delta(G)$, quantifying the relative change in total FLOPs when applying Group-\acp{sae} instead of per-layer \acp{sae}, is defined as:

\textcolor{black}{\begin{equation}
\label{eqn:flops-saving}
\Delta(G) = 1 - \frac{A + G\,T}{A + L\,T}.
\end{equation}
By definition, if $G = L$, then $\Delta(G) = 0$, meaning no savings. Conversely, as $G$ decreases, savings increase, reaching a maximum of ${(LT - T) / (A + LT)}$ when $G = 1$, which approaches $(L-1)/L$ as the ratio $T/A$ increases.}

Since our method does not alter either $A$ or $T$, the efficiency gains of Group-\acp{sae} are primarily determined by the $G/L$ ratio. 

\section{Experiments}
\label{sec:exps}

Our work is primarily focused on addressing the following research questions:
\begin{itemize}
    \item[\textbf{Q1}] Do SAEs trained on groups of layers activations maintain reconstruction quality and downstream performance?
    \item[\textbf{Q2}] Does selecting the number of groups \(G\) based on the Average Maximum Angular Distance ({AMAD}) ensure an optimal balance between computational efficiency and model performance?
    \item[\textbf{Q3}] How do Group-SAEs affect the interpretability of the SAE latent representations?
\end{itemize}
To address these questions, we compare the performance of standard SAEs and Group-SAEs across a range of metrics and alternative grouping strategies. 

\label{sec:experiments}
\subsection{Experimental setting}
We denote \saei{l} as the baseline \ac{sae} trained to reconstruct the activations of layer $l$. For every \( g = 1,\dots, G \), with \(G \in \{1, \dots, L-1\}\) and $L$ being the number of layers of a model, let \([g_G]\) represent the set of layers belonging to the \( g \)-th group within the partition of G groups. We then define \saeki as the \ac{sae} trained to reconstruct the activations for all layers in \([g_G]\). To ensure a fair comparison with baselines, we allocate 1 billion training tokens for each \saei{l} and \saeki. For baseline \acp{sae}, activations are always taken from a single fixed layer. In contrast, for Group-\acp{sae}, activations are drawn from a randomly selected layer within the set $[g_G]$. In this way, we ensure that each Group- and baseline \acp{sae} process exactly 1 billion tokens and activations.

\begin{figure*}[ht]
    \begin{subfigure}{.48\textwidth}
        \centering
        \includegraphics[width=\linewidth]{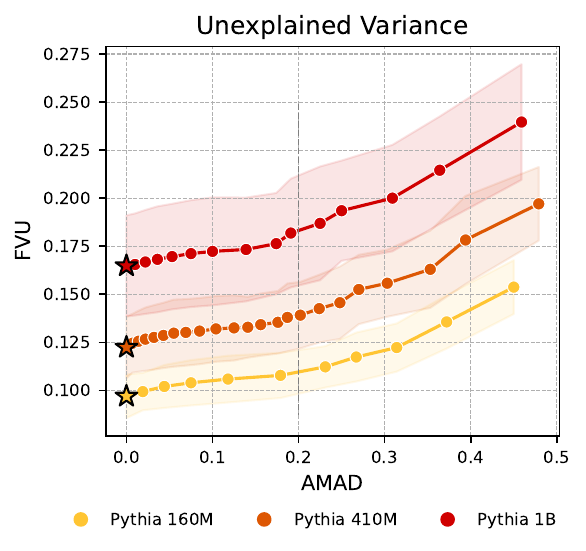}
    \end{subfigure}
    \begin{subfigure}{.48\textwidth}
        \centering
        \includegraphics[width=\linewidth]{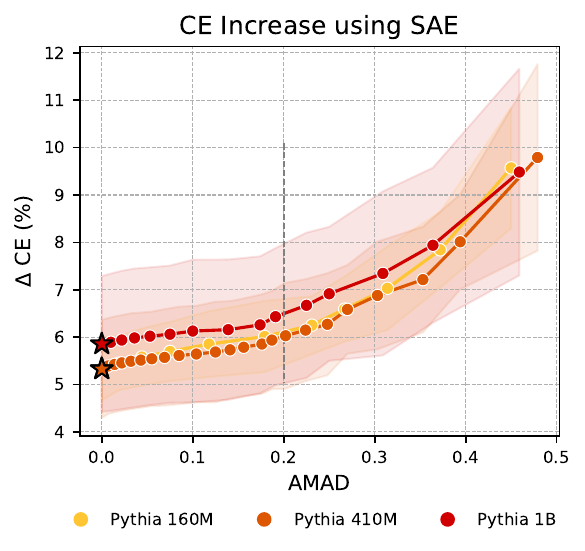}
    \end{subfigure}
    \caption{(Left) FVU and (Right) $\Delta\text{CE}(\%)$ over $\text{{AMAD}}(G)$ for every $G \in \{1,\dots,L-1\}$. The highlighted star markers represent the baseline \acp{sae} (i.e., with no grouping), while the other points correspond to Group-\acp{sae}, ordered from left to right by increasing {AMAD}, which reflects a decrease in the number of groups. The shaded area indicates one std.}
    \label{fig:amd_vs_metric}
\end{figure*}

\paragraph{Models, Dataset and Hyperparameters}
\label{subsec:hyperparams}
Following~\citet{lawson2024mlsae}, we train both \acp{sae} and Group-\acp{sae} with the Fraction of Variance Unexplained (FVU) as reconstruction loss. Defined as
\begin{equation}
    \text{FVU}(\mathbf x)=\frac{\|\mathbf{x} - \widehat{\mathbf x}\|_2^2}{\text{Var}(\mathbf x)},
    \label{eqn:fvu_loss}
\end{equation}
we prefer it to standard MSE loss as it accounts for the different magnitudes of activations coming from different layers of the model. We employ Top-$K$ activation\footnote{The Top-$K$ activation function is directly applied on the features obtained with Equation\ref{eqn:encode}, where $\sigma = \text{Top-}K \circ \text{ReLU}$.} with $K=128$ and expansion factor of $c=16$ on the residual stream after the MLP contribution of three models of varying sizes from the Pythia family~\cite{biderman2023pythia}: Pythia 160M, Pythia 410M, and Pythia 1B. 

\textcolor{black}{We follow established practice in \ac{sae} training \citep{bricken2023monosemanticity, gao2024topk, rajamanoharan2024jumprelu} and use the same pre-training dataset as the models themselves. In particular, we sample 1 billion tokens from the Pile dataset~\cite{pile} and process them with a context size of 1024.}

For each model, we compute all partitions ${G \in \{1,\dots, L-1\}}$ and train a Group-SAE for all groups of layers in them. We exclude the last layer from all partitions because it resides in the unembedding space and, based on our empirical findings, consistently exhibits a distinct reconstruction error pattern. As a result, it requires a separate \ac{sae}. Additionally, we compare our grouping strategy with two \textbf{baseline techniques} aimed to reduce the computational cost of training SAEs: (1) training Group SAEs on evenly spaced groups, and (2) training smaller SAEs on all layers. Hyperparameters for all the experiments and training details can be found in Appendix~\ref{app:hyperparameters} and~\ref{app:training} respectively.

\paragraph{Evaluation.} \textcolor{black}{As in previous work \cite{huben2024sparse, gao2024topk}, we evaluate quality of trained SAE} across three key areas: reconstruction, downstream, and interpretability.

For both reconstruction and downstream evaluations, we use a subset of the Pile dataset (distinct from the training set) comprising 1 million tokens.

For reconstruction, we compare each \saeki with its corresponding baseline \saei{l} for every layer \(l \in [g_G]\). We report the average Fraction of Variance Unexplained (FVU, Equation~\ref{eqn:fvu_loss}) as our reconstruction metric. 

To evaluate downstream performance, we measure the effect of replacing a layer's activation with its SAE reconstruction on the next-token prediction. Specifically, we compute the average relative change in next-token Cross-Entropy:
\begin{equation}
\Delta\text{CE} = \frac{\text{CE}(\text{M}(P \mid \mathbf{x}^l \leftarrow \widehat{\mathbf{x}}^l)) - \text{CE}(\text{M}(P))}{\text{CE}(\text{M}(P))},
\end{equation}
where \(\text{M}\) denotes the model, \(P\) is the input prompt, and \(\text{M}(P \mid \mathbf{x}^l \leftarrow \widehat{\mathbf{x}}^l)\) indicates the model output when the true activation \(\mathbf{x}^l\) at layer \(l\) is replaced with the SAE reconstruction \(\widehat{\mathbf{x}}^l\).

\begin{table*}[ht]
\centering
\caption{FVU and $\Delta\text{CE}$ for different approaches across model sizes. Our proposed grouping strategy based on the AMAD achieves lower FVU and $\Delta\text{CE}$ compared to the baselines: Group SAEs with evenly spaced groups and smaller SAEs trained on all layers. Note that both the \textit{Evenly Spaced} and \textit{Smaller SAEs} strategies have the same number of training FLOPs as our AMAD-based grouping strategy. For each entry, the value in \% shown to the right indicates the relative improvement over \textit{Smaller SAEs (All layers)} baseline (positive = better).}
\begingroup
\setlength{\tabcolsep}{4pt}
\renewcommand{\arraystretch}{1.1}
\resizebox{\textwidth}{!}{%
\begin{tabular}{lcccccc}
\toprule
\multirow{2}{*}{\textbf{Approach}} &
\multicolumn{2}{c}{\textbf{Pythia-160M}} &
\multicolumn{2}{c}{\textbf{Pythia-410M}} &
\multicolumn{2}{c}{\textbf{Pythia-1B}} \\
\cmidrule(lr){2-3} \cmidrule(lr){4-5} \cmidrule(lr){6-7}
 & \textbf{FVU} & \textbf{$\Delta\text{CE}_{\%}$} &
   \textbf{FVU} & \textbf{$\Delta\text{CE}_{\%}$} &
   \textbf{FVU} & \textbf{$\Delta\text{CE}_{\%}$} \\
\midrule
\textbf{Group SAEs (AMAD with $\widehat{G}$ groups)} &
$\mathbf{0.108}$ {\scriptsize\,(+6.1\%)} & 6.01 {\scriptsize\,(+18.5\%)} &
$\mathbf{0.138}$ {\scriptsize\,(+5.5\%)} & \textbf{5.94} {\scriptsize\,(+16.3\%)} &
$\mathbf{0.182}$ {\scriptsize\,(+3.2\%)} & \textbf{6.43} {\scriptsize\,(+20.6\%)} \\
Group SAEs (Evenly spaced with $\widehat{G}$ groups) &
0.114 {\scriptsize\,(+0.9\%)} & \textbf{5.40} {\scriptsize\,(+26.7\%)} &
0.145 {\scriptsize\,(+0.7\%)} & 6.01 {\scriptsize\,(+15.4\%)} &
0.189 {\scriptsize\,($-0.5$\%)} & 6.63 {\scriptsize\,(+18.1\%)} \\
\midrule
Smaller SAEs (All layers) &
0.115 {\scriptsize\,(+0.0\%)} & 7.37 {\scriptsize\,(+0.0\%)} & 0.146 {\scriptsize\,(+0.0\%)} & 7.10 {\scriptsize\,(+0.0\%)} & 0.188 {\scriptsize\,(+0.0\%)} & 8.10 {\scriptsize\,(+0.0\%)} \\
\bottomrule
\end{tabular}
}%
\endgroup
\label{tab:ablation_study_with_diffs}
\end{table*}

For interpretability, we adopt the automated pipeline proposed by \citet{paulo2024automatically}. First, an \textit{explainer} language model (LM) generates natural language explanations of the SAE latent representations. Then, a separate \textit{scorer} LM evaluates these explanations. In our experiments, both the explainer and scorer are implemented using \texttt{gemini-2.0-flash-001}\footnote{\url{https://deepmind.google/technologies/gemini/flash/}}. Specifically, for each SAE, we randomly sample 64 features and cache their latent activations over a 10M token sample from the Pile. For each latent, the explainer is shown 20 distinct examples, 10 activating the latent and 10 sampled randomly, each consisting of 32 tokens. Two binary scoring strategies are employed:
\begin{itemize}
    \item \textit{Detection}: A language model determines whether a given sequence activates an SAE latent according to the provided explanation.
    \item \textit{Fuzzing}: Activating tokens are marked within each example, and a language model is prompted to assess whether the marked sentences are correctly identified.
\end{itemize}
Figure~\ref{fig:auto-interp-examples} in Appendix~\ref{app:auto-interp} shows a sentence example for each strategy.
For every metric (FVU, $\Delta\text{CE}$ and Detection/Fuzzing) and for each \(G \in \{1, \dots, L-1\}\), we first compute all metrics at the layer level, then aggregate the results for each partition \(g\) within \(G\) by computing the mean and the standard deviation weighted by the number of layers in that partition.

\subsection{Results}
\label{subsec:res}
In the following paragraphs, we aim to empirically answer the research questions outlined in Section~\ref{sec:exps}.

\begin{figure*}[ht!]
    \begin{subfigure}{.48\textwidth}
        \centering
        \includegraphics[width=\linewidth]{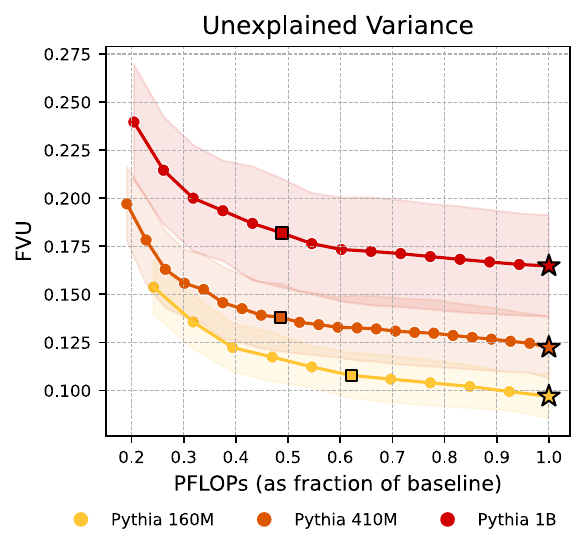}
    \end{subfigure}\hspace*{\fill}
    \begin{subfigure}{.48\textwidth}
        \centering
        \includegraphics[width=\linewidth]{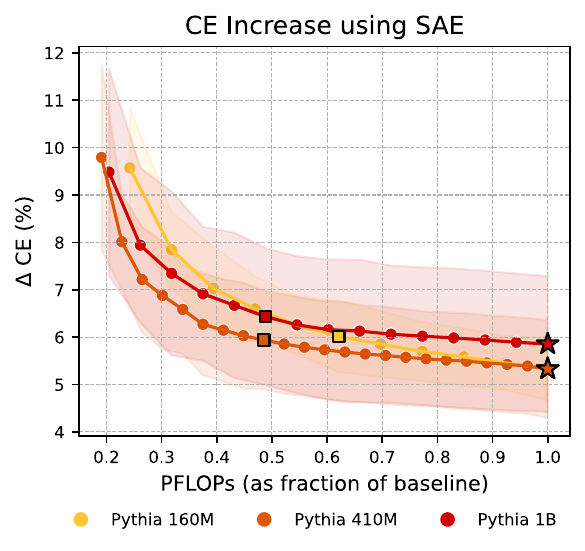}
    \end{subfigure}
    \caption{(Left) FVU and (Right) $\Delta\text{CE}(\%)$ over the fraction of training PFLOPs with respect to the baseline. The highlighted star markers represent the baseline \acp{sae} (i.e., with no grouping), while the other points correspond to Group-\acp{sae}, ordered from right to left by decreasing PFLOPs, which reflects a decrease in the number of groups. The highlighted square markers represent the Group-\acp{sae} with a number of groups ${\widehat{G} = \min\{G \mid \text{{AMAD}}(G) < 0.2\}}$.}
    \label{fig:rel_flops_vs_metric}
\end{figure*}

\paragraph{Q1: What is the impact of grouping layers (Group-SAEs) on reconstruction quality and downstream task performance?}
In Figure~\ref{fig:amd_vs_metric}, we plot the average FVU and the cross-entropy difference (\(\Delta\text{CE}\)) as functions of the {AMAD} for different group configurations. The highlighted star markers represent the baseline models (i.e., with no grouping), while the other points correspond to grouped models. The points are ordered from left to right by increasing {AMAD}, which reflects a decrease in the number of groups, with \(G\) ranging from \(L-1\) down to \(1\). From Figure~\ref{fig:amd_vs_metric}, a \textit{notable turning point} emerges around \(\text{{AMAD}}(G)\approx 0.2\): increasing {AMAD} beyond this threshold leads to a more rapid loss in performance. In particular, training a single SAE on all the model layers ($G=1$), although achieving the best computational saving, also incurs the worst reconstruction and downstream performance. 

To further validate our method, in Appendix~\ref{app:f_sim}, we inspect the quality of features learned by Group-\acp{sae} by measuring their similarity to the features learned by Baseline-SAEs. As expected, for each baseline \saei{l}, we found average similarity to peak with the Group-\ac{sae} trained on a group containing~$l$. Moreover, Appendix~\ref{app:f_dist} analyzes how a Group-\ac{sae}’s features distribute across the activations of layers in its group: consistent with the analysis of \citet{lindsey2024crosscoders}, individual features typically peak at a specific layer yet exhibit substantial spread to adjacent layers. This pattern supports our core hypothesis that while features may anchor to particular layers, they remain relevant across neighboring ones, enabling a single \ac{sae} to effectively reconstruct activations from multiple, similar layers.

\paragraph{Q2: Does selecting the number of groups \(G\) based on the {AMAD} ensure an optimal balance between computational efficiency and model performance?}
Motivated by the insights from the previous paragraph, the optimal $G$ is chosen as ${\widehat{G} = \min\{G \mid \text{{AMAD}}(G) < 0.2\}}$. In Figure~\ref{fig:rel_flops_vs_metric} we show both FVU and \(\Delta\text{CE}\) plotted against the fraction of PFLOPs relative to the baseline. Again, star markers denote baseline SAEs, whereas circles represent Group-SAEs. Here, moving from right to left indicates reducing PFLOPs (i.e., training fewer SAEs overall). The points are ordered from right to left by decreasing PFLOPs, which reflects a decrease in the number of groups, from \(L-1\) down to \(1\). The highlighted square markers correspond to Group-SAEs with \(\widehat{G}\) groups; they substantially reduce training costs up to more than 50\% with only a moderate performance penalty: ${\text{FVU}(\bigstar) - \text{FVU}(\blacksquare) \approx -0.015}$ and ${\Delta\text{CE}_{\%}(\bigstar) - \Delta\text{CE}_{\%}(\blacksquare) \approx -0.62}$ for all three evaluated models.

\begin{figure*}[ht]
    \centering
    \begin{subfigure}{0.48\textwidth}
        \centering
        \includegraphics[width=\linewidth]{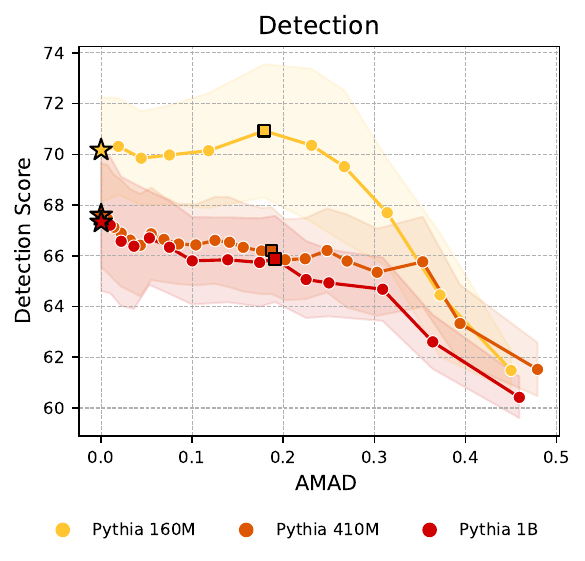}
    \end{subfigure}
    \hfill
    \begin{subfigure}{0.48\textwidth}
        \centering
        \includegraphics[width=\linewidth]{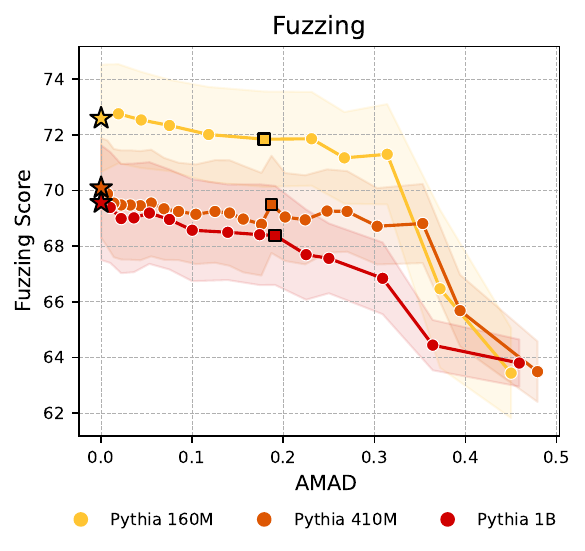}
    \end{subfigure}
    \caption{Auto-Interpretability scores following the automated pipeline defined by~\cite{paulo2024auto-interp} over $\text{{AMAD}}(G)$ for every $G \in \{1,\dots,L-1\}$. The highlighted star markers represent the baseline \acp{sae} (i.e., with no grouping), while the other points correspond to Group-\acp{sae}, ordered from left to right by increasing {AMAD}, which reflects a decrease in the number of groups. The highlighted square markers represent the Group-\acp{sae} with a number of groups ${\widehat{G} = \min\{G \mid \text{{AMAD}}(G) < 0.2\}}$. (Left) Detection and (Right) Fuzzing scores, as defined in the Evaluation paragraph of Section~\ref{sec:experiments}.}
    \label{fig:interp_all}
\end{figure*}

To ensure that our grouping strategy and the selection of $\widehat{G}$ based on AMAD offer an effective trade-off between computational efficiency and performance, we compare them against two baselines: 1) \textit{Evenly Spaced Group SAEs}: Group SAEs trained such that each partition contains nearly equal numbers of layers; 2) \textit{Smaller SAEs}: A separate, smaller SAE is trained for each layer. All methods are adjusted to incur equal computational costs\footnote{For Evenly Spaced Group SAEs, we use the same number of groups $\widehat{G}$; for Smaller SAEs, we set the expansion factor as $c' = c \cdot \widehat{G}/T$, matching the FLOPs of a Group-SAE with $\widehat{G}$ groups.}. Results in Table~\ref{tab:ablation_study_with_diffs} shows that the proposed method outperforms the two additional baselines across nearly all models and evaluation metrics, with only a single exception observed in the case of Pythia-160M. Importantly, this exception does not arise from the idea of grouping layers but from the chosen grouping strategy. Indeed, our method consistently outperforms the standard per-layer approach with smaller Standard SAEs. We observe that these advantages are particularly noticeable for the $\Delta$CE metric, related to the downstream performance. Additionally, Table~\ref{tab:flops_comparison} presents the computational costs and savings, as defined in Eq.~\ref{eqn:flops-saving} of Section~\ref{sec:complexity}, of Group-\acp{sae} compared to the baselines when the optimal number of groups $G$ is selected as $\widehat{G}$.

\begin{table}[!t]
\centering
\caption{Comparison of FLOPs ($10^{18}$) required for caching activations and training Baseline and Group \acp{sae} on 1B tokens, covering all layers with an expansion factor of 16 and ${\widehat G = \min \{G \mid \text{{AMAD}}(G) < 0.2\}}$.}
\label{tab:flops_comparison}
\begin{tabular}{l@{\hskip 5pt}c@{\hskip 5pt}c@{\hskip 5pt}c@{\hskip 5pt}c}
\toprule
\textbf{Model} & $\widehat{\textbf{G}}$ & \textbf{A+LT}  & \textbf{A+{$\widehat{\text{G}}$}T} & $\mathbf{\Delta}_\%(\widehat{\mathbf{G}})$ \\
\midrule
Pythia 160M & 6 & $1.34$  & $\mathbf{0.77}$ & $+42.5\%$ \\
Pythia 410M & 9 & $4.73$  & $\mathbf{2.21}$ & $+53.3\%$ \\
Pythia 1B   & 6 & $12.48$ & $\mathbf{5.77}$ & $+53.7\%$ \\
\bottomrule
\end{tabular}
\end{table}

\paragraph{Q3: How do Group-SAEs affect the interpretability of the SAE latent representations?}
\label{sec:interp}
\textcolor{black}{To answer, we employ the auto-interpretability pipeline proposed by \cite{paulo2024auto-interp}.}
For each SAE latent, first, an \textit{explainer} Language Model is asked to propose a natural language explanation of it given both activating and non-activating examples. Then, given the explanation, a \textit{scorer} Language Model is tasked with predicting the set of sentences that should activate the target latent (\textit{detection}) and the sentences containing highlighted tokens that activate the target latent (\textit{fuzzing}).
In Figure~\ref{fig:interp_all} we plot both the detection and fuzzing scores for all the evaluated models. In the figures, square markers denote Group-SAEs with $\widehat{G}$ groups, while star markers indicate the baseline SAEs. We observe a similar trend as in reconstruction and downstream evaluations: detection and fuzzing scores improve more rapidly as $\text{AMAD}(G)$ decreases—provided it remains above the turning point—after which the scores plateau at an approximately constant level. This result further validates our selection of $\widehat{G}$ based on AMAD, suggesting that the interpretability of features in the baseline and Group-SAEs differs only marginally.

\section{Conclusion}
\label{sec:conclusion}
This work introduces a novel approach to efficiently train \acp{sae} for \acp{llm} by clustering layers based on their angular distance and training a single SAE for each group. Through this method, we achieved up to a 50\% reduction in training costs without compromising reconstruction quality or performance on downstream tasks. The results demonstrate that activations from adjacent layers in LLMs share common features, enabling effective reconstruction with fewer \acp{sae}.

Our findings also show that the \acp{sae} trained on grouped layers perform comparably to layer-specific \acp{sae} in terms of reconstruction and downstream metrics. Furthermore, the automated interpretability evaluations confirmed the interpretability of the features learned by our \acp{sae}, underscoring their utility in disentangling neural activations.

The methodology proposed in this paper opens avenues for more scalable interpretability tools, facilitating deeper analysis of LLMs as they grow in size. Future work will focus on further optimizing the number of layer groups and scaling the approach to even larger models.

\section*{Limitations}
\label{sec:limitations}
Although we evaluated our approach across various groups and model sizes, our primary focus here is on experiments using a fixed expansion factor of $c=16$ and TopK as activation function. Even if we don't expect the choices of these hyper-parameters to influence the results of this work, we left investigations of this phenomenon for future work. 

\textcolor{black}{We also limit the scope of our study to models from the Pythia family trained on the Pile dataset. While using the pre-training dataset for SAE training is standard practice \citep{bricken2023monosemanticity, gao2024topk}, evaluating on additional datasets could provide stronger evidence of generality. We leave such cross-dataset evaluation to future work. Furthermore, architectural and training differences across model families may influence the behavior of Group-SAEs.} We defer a comprehensive cross-model analysis to future research. Exploring the generality of our findings across diverse architectures, such as Gemma, LLaMA, Qwen, or Mistral, is an important next step. Finally, our interpretability evaluation remains limited, primarily due to the high economic cost of annotating large numbers of features. While we observe promising patterns, a more comprehensive and systematic interpretability analysis is left for future work.

\section*{Reproducibility statement}
To support the replication of our empirical findings on training \acp{sae} via layer groups and to enable further research on understanding their inner works, \textcolor{black}{we release the code and \acp{sae} used in this study~\footnote{https://github.com/ghidav/group-sae}.}


\textcolor{black}{
\section*{Acknowledgements}
The work has received funding from the European Union’s Horizon Europe research and innovation programme under grant agreements No. 101189771 (DataPACT) and No. 101070284 (enRichMyData), and the Italian PRIN project Discount Quality for Responsible Data Science (202248FWFS). Additionally, we acknowledge and thank Nscale for providing the compute resources (8 AMD Mi250x GPUs) used for all SAE training and most evaluations in this paper. We are especially grateful to Karl Havard for leading this partnership, Konstantinos Mouzakitis for his technical assistance, Brian Dervan for structuring our collaboration, and the entire Nscale team for their support.
}

\bibliography{anthology, custom}

\newpage
\onecolumn
\appendix

\section{Hyperparameters}
\label{app:hyperparameters}
We train both \acp{sae} and Group-\acp{sae} using Top-$K$ activation\footnote{The Top-$K$ activation function is directly applied on the features obtained with Equation\ref{eqn:encode}, where $\sigma = \text{Top-}K \circ \text{ReLU}$.} with $K=128$ and expansion factor of $c=16$ on the residual stream after the MLP contribution of three models of varying sizes from the Pythia family~\cite{biderman2023pythia}: Pythia 160M, Pythia 410M, and Pythia 1B. 
To train all the \acp{sae}, we sample 1 billion tokens from the Pile dataset~\cite{pile} and process them with a context size of 1024.
We use Adam optimizer~\cite{kingma2017adam} with default $\beta$ parameters and set the learning rate equal to $2\text{e-}4 / \sqrt{(m / 2^{14})}$ as specified in \citet{gao2024topk}. We use a batch size of $131072, 65536\text{ and } 32768$ for the three models, respectively, to maximize computational usage. Following~\cite{bricken2023monosemanticity} we constrain the decoder columns (i.e. the feature directions) to have unit norm. Additionally, we normalize the activations to have mean squared $\ell_2$ norm of 1 during \ac{sae} training, as specified in~\cite{rajamanoharan2024jumprelu}, by first estimating the norm scaling factor over 5 million tokens of our train set.

\begin{table*}[h]
\centering
\caption{Pythia model details.}
  \begin{tabular}{lcccc}
    \toprule
    Pythia model & Non-Embedding Params & Layers & Model Dim & Heads \\
    \midrule
    160M  & 85,056,000  & 12 & 768  & 12 \\
    410M  & 302,311,424 & 24 & 1024 & 16 \\
    1.0B  & 805,736,448 & 16 & 2048 & 8  \\
    \bottomrule
  \end{tabular}%
\label{tab:pythia_models}
\end{table*}

\begin{table*}[h]
    \caption{Training and fine-tuning hyperparameters}
    \centering
    \begin{tabular}{ll}
        \toprule
        \textbf{Hyperparameter} & \textbf{Value}\\
        \midrule[\heavyrulewidth]
        c & 16\\
        Top-$K$ K & 128\\
        $\alpha_{\text{aux}}$ & $1/32$ \\
        Hook name & resid-post\\
        \multirow{3}{*}{Batch size} 
          & 131'072 (Pythia-160M) \\ 
          & 65'536 (Pythia-410M) \\ 
          & 32'768 (Pythia-1B) \\
        Adam $(\beta_1, \beta_2)$ & $(0.9, 0.999)$\\
        Context size & 1024\\
        $\text{lr}$ & $2\text{e-}4 / \sqrt{(m / 2^{14})}$\\
        $\text{lr}$ scheduler & constant\\
        Dead latents threshold & 10M \\
        $\#$ tokens (Train) & 1B\\
        Checkpoint freq & 100K\\
        Decoder column normalization & Yes\\
        Activation normalization & Mean squared $\ell_2$ norm equal to 1 during SAE training\\
        FP precision & 32\\
        Prepend BOS token & No\\
        \bottomrule
    \end{tabular}
    \label{tab:hyperparams}
\end{table*}

The experiments were carried out on a cluster of 8 AMD MI250X. The longest experimental run took approximately 24 hours.
Our experiments were carried out using \texttt{PyTorch} \cite{NEURIPS2019_bdbca288} and the \texttt{sparsify} library.\footnote{\url{https://github.com/EleutherAI/sparsify}}
We performed our data analysis using \texttt{NumPy} \citep{harris2020array} and \texttt{Pandas} \citep{mckinney-proc-scipy-2010}. Our figures were made using \texttt{Matplotlib} \citep{Hunter:2007} and \texttt{Seaborn} \citep{Waskom2021}.

\clearpage
\newpage
\section{SAEs Training Details}
\label{app:training}
Following \citet{lawson2024mlsae}, given $\mathbf{X},\widehat{\mathbf{X}}\in\mathbb R^{B\times n}$ being the input activation batch and its \ac{sae} reconstruction, respectively, we train our \acp{sae} with the following loss:
\begin{equation}
    \mathcal L(\mathbf X)=\text{FVU}(\mathbf{X},\,\widehat{\mathbf X})+\alpha_\text{aux}\cdot\text{AuxK}(\mathbf{X},\,\widehat{\mathbf X})
    \label{eqn:topk-loss}
\end{equation}
The first term of the loss is the Fraction of Variance Unexplained, or:
\begin{equation}
    \text{FVU}(\mathbf{X}, \mathbf{\widehat{X}}) = \frac{\|\mathbf{X}-\widehat{\mathbf X}\|_F}{\|\mathbf{X} - \overline{\mathbf X}\|_F}
\label{eqn:fvu}
\end{equation}
where $\|\cdot\|_F$ is the Frobenius norm and $\overline{\mathbf X}=\frac 1B \mathbf{1}_B \mathbf{1}_B^\top\mathbf{X}$ is a matrix where each row corresponds to the mean of $\mathbf X$ along the batch dimension. The second term of the loss is an auxiliary loss to prevent the formation of dead latents during training and is defined as:
\begin{equation}
    \text{AuxK}(\mathbf{X}, \mathbf{\widehat{X}}) = \frac{\|\mathbf{E}-\widehat{\mathbf E}\|_F}{\|\mathbf{X} - \overline{\mathbf X}\|_F}
\label{eqn:auxk}
\end{equation}
Here, $\mathbf{E}=\mathbf{X}-\widehat{\mathbf X}$ is the reconstruction error of the main model, and $\widehat{\mathbf E}$ is its reconstruction using the top-$\text{K}_{\text{aux}}$ dead latents. A dead latent $\mathbf{f}_i(\mathbf{x})$ is a latent that didn't fire, i.e. ${\mathbf{f}_i(\mathbf{x}) = 0}$, for a predefined number of {tokens} (10M in our experiments). Following \citet{gao2024topk}, we choose $K_{\text{aux}}$ as the minimum between the number of dead latents and $m/2$, and $\alpha = 1/32$.

To ensure a fair comparison with baselines, we allocate 1 billion training tokens for each \saei{l} and \saeki. For baseline \acp{sae}, activations are always taken from a single fixed layer. In contrast, for Group-\acp{sae}, activations are drawn from a randomly selected layer within the set $[g_G]$. In this way, we ensure that each Group- and baseline \acp{sae} process exactly 1 billion tokens and activations.

\clearpage
\newpage
\section{Angular Distances and Layers Groups}
\label{app:ang-dist}
We use the same angular distance formulation of~\citet{gromov2024unreasonable}:
\begin{equation}
\label{eq:angular_dist}
d_\theta\left(\mathbf{x}^i, \mathbf{x}^j\right) = \frac{1}{\pi} \arccos{\left(\frac{\mathbf{x}^i \cdot \mathbf{x}^j}{\norm{\mathbf{x}^i}_2\norm{\mathbf{x}^j}_2}\right)}
\end{equation}
for every $i,j \in \{1,...,L\}$, where $\mathbf{x}^l$ are the $l$-th residual stream activations after the MLP's contribution.

\begin{figure*}[h]
    \centering
    \includegraphics[width=0.4\linewidth]{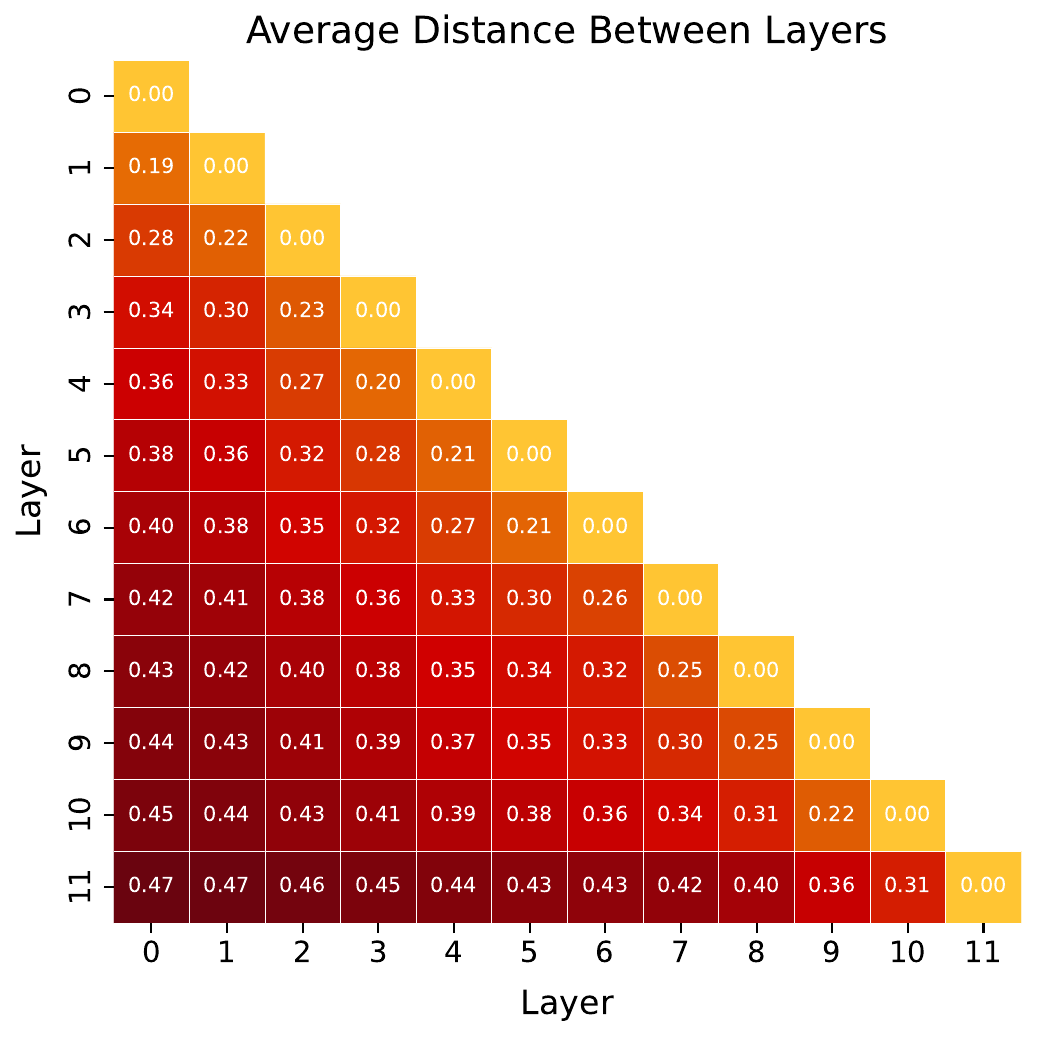}
    \caption{Average angular distance between all layers of the Pythia-160M model, as defined in Equation~\ref{eq:angular_dist}. The angular distances are computed over 10M tokens from the training dataset. The angular distances are bounded in $[0,1]$, where an angular distance equal to $0$ means equal activations, $0.5$ means activations are perpendicular and an angular distance of $1$ means that the activations point in opposite directions.}
    \label{fig:pythia_160m_angular_distances}
\end{figure*}

\begin{figure*}[h]
    \centering
    \includegraphics[width=0.5\linewidth]{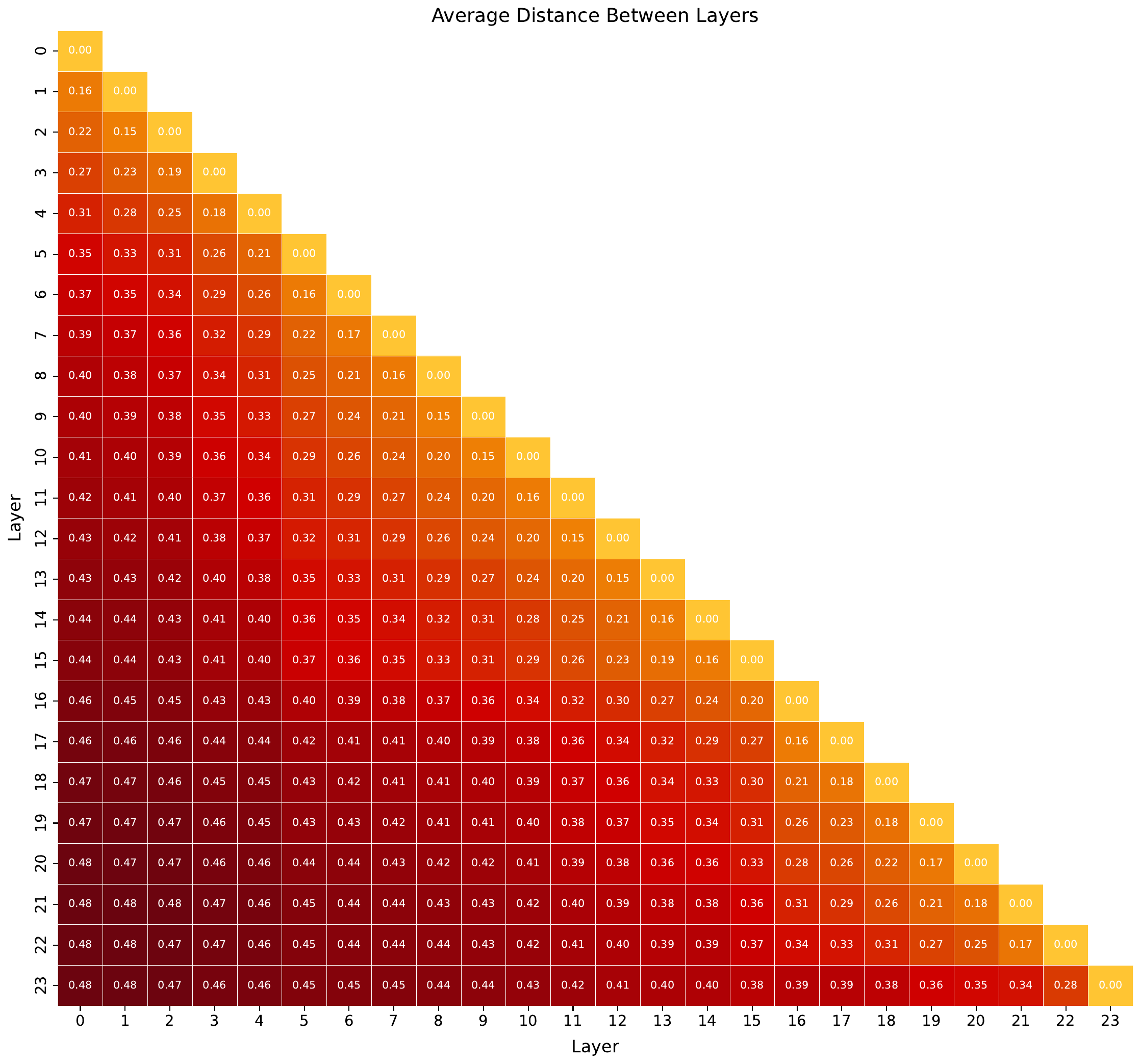}
    \caption{Average angular distance between all layers of the Pythia-410M model, as defined in Equation~\ref{eq:angular_dist}. The angular distances are computed over 10M tokens from the training dataset. The angular distances are bounded in $[0,1]$, where an angular distance equal to $0$ means equal activations, $0.5$ means activations are perpendicular and an angular distance of $1$ means that the activations point in opposite directions.}
    \label{fig:pythia_410m_angular_distances}
\end{figure*}

\begin{figure*}[h]
    \centering
    \includegraphics[width=0.55\linewidth]{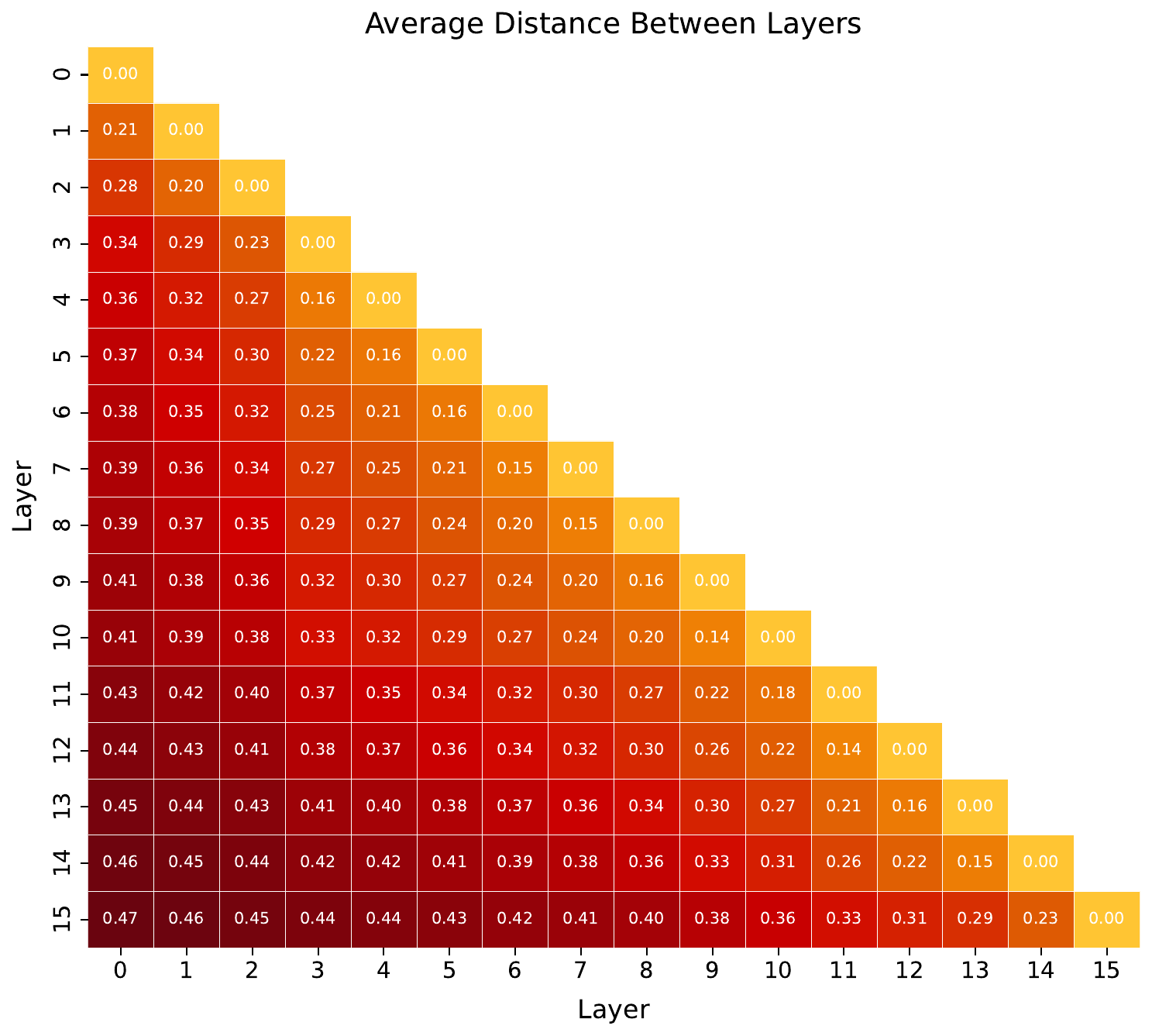}
    \caption{Average angular distance between all layers of the Pythia-1B model, as defined in Equation~\ref{eq:angular_dist}. The angular distances are computed over 10M tokens from the training dataset and are bounded in $[0,1]$. An angular distance equal to $0$ means equal activations, $0.5$ means activations are perpendicular and an angular distance of $1$ means that the activations point in opposite directions.}
    \label{fig:pythia_1b_angular_distances}
\end{figure*}

Figures~\ref{fig:pythia_160m_angular_distances}–\ref{fig:pythia_1b_angular_distances} visualize the \emph{pairwise} average angular distances between residual-stream activations across all layers for each Pythia model (computed on 10M training tokens). Values are scaled to $[0,1]$ (0 = identical directions, $0.5$ = orthogonal, 1 = opposite), with block structure revealing contiguous regions of high similarity that motivate layer grouping. Tables~\ref{tab:pythia-160m-groups}–\ref{tab:pythia-1b-groups} then report the grouping solutions as we vary the number of groups $G$ (up to $L-1$): the \textit{Groups} row lists, for each layer index, the assigned group ID, and the accompanying \textit{AMAD} value (Average Maximum Angular Distance, Eq.~\ref{eqn:amad}) summarizes within-group compactness. As $G$ increases, AMAD typically decreases, reflecting finer partitions that better capture the block-diagonal structure observed in the distance matrices.

\begin{table}[h]
\centering
\resizebox{.3\textwidth}{!}{%
\begin{tabular}{lcl}
\toprule
\textbf{G} & \textbf{Groups} & \textbf{{AMAD}} \\
\midrule
1  & \small{0, 0, 0, 0, 0, 0, 0, 0, 0, 0, 0}                & 0.450  \\
2  & \small{0, 0, 0, 0, 0, 0, 0, 1, 1, 1, 1}                & 0.372 \\
3  & \small{2, 2, 2, 1, 1, 1, 1, 0, 0, 0, 0}                & 0.314 \\
4  & \small{2, 2, 2, 0, 0, 0, 0, 1, 1, 3, 3}                & 0.267 \\
5  & \small{0, 0, 0, 4, 4, 2, 2, 1, 1, 3, 3}                & 0.231 \\
6  & \small{3, 3, 5, 4, 4, 2, 2, 0, 0, 1, 1}                & 0.179 \\
7  & \small{3, 3, 5, 1, 1, 2, 2, 6, 4, 0, 0}                & 0.118 \\
8  & \small{3, 3, 5, 1, 1, 0, 0, 6, 4, 7, 2}                & 0.075 \\
9  & \small{1, 1, 5, 0, 0, 8, 7, 6, 4, 3, 2}                & 0.044 \\
10 & \small{0, 0, 5, 9, 7, 8, 3, 6, 4, 1, 2}                & 0.019 \\
\bottomrule
\end{tabular}%
}
\caption{Layer groups for every $G$ up to $L-1$ for Pythia-160M}
\label{tab:pythia-160m-groups}
\end{table}

\begin{table}[h]
\centering
\resizebox{.6\textwidth}{!}{%
\begin{tabular}{lcl}
\toprule
\textbf{G} & \textbf{Groups} & \textbf{{AMAD}} \\
\midrule
1  & \small{0, 0, 0, 0, 0, 0, 0, 0, 0, 0, 0, 0, 0, 0, 0, 0, 0, 0, 0, 0, 0, 0, 0} & 0.479 \\
2  & \small{0, 0, 0, 0, 0, 0, 0, 0, 0, 0, 0, 0, 0, 0, 0, 0, 1, 1, 1, 1, 1, 1, 1} & 0.394 \\
3  & \small{2, 2, 2, 2, 2, 2, 2, 0, 0, 0, 0, 0, 0, 0, 0, 0, 1, 1, 1, 1, 1, 1, 1} & 0.353 \\
4  & \small{2, 2, 2, 2, 2, 2, 2, 3, 3, 3, 3, 1, 1, 1, 1, 1, 0, 0, 0, 0, 0, 0, 0} & 0.303 \\
5  & \small{0, 0, 0, 0, 0, 0, 0, 3, 3, 3, 3, 1, 1, 1, 1, 1, 4, 4, 4, 2, 2, 2, 2} & 0.270 \\
6  & \small{5, 5, 5, 1, 1, 1, 1, 3, 3, 3, 3, 0, 0, 0, 0, 0, 4, 4, 4, 2, 2, 2, 2} & 0.248 \\
7  & \small{5, 5, 5, 0, 0, 0, 0, 1, 1, 1, 1, 6, 6, 3, 3, 3, 4, 4, 4, 2, 2, 2, 2} & 0.224 \\
8  & \small{2, 2, 2, 5, 5, 7, 7, 1, 1, 1, 1, 6, 6, 3, 3, 3, 4, 4, 4, 0, 0, 0, 0} & 0.202 \\
9  & \small{2, 2, 2, 5, 5, 7, 7, 0, 0, 0, 0, 6, 6, 3, 3, 3, 1, 1, 1, 8, 8, 4, 4} & 0.187 \\
10 & \small{2, 2, 2, 5, 5, 7, 7, 8, 8, 9, 9, 6, 6, 1, 1, 1, 0, 0, 0, 3, 3, 4, 4} & 0.176 \\
11 & \small{2, 2, 2, 5, 5, 7, 7, 8, 8, 9, 9, 6, 6, 0, 0, 0, 10, 4, 4, 3, 3, 1, 1} & 0.156 \\
12 & \small{0, 0, 0, 2, 2, 7, 7, 8, 8, 9, 9, 6, 6, 11, 5, 5, 10, 4, 4, 3, 3, 1, 1} & 0.141 \\
13 & \small{12, 9, 9, 2, 2, 7, 7, 8, 8, 4, 4, 6, 6, 11, 5, 5, 10, 1, 1, 3, 3, 0, 0} & 0.125 \\
14 & \small{12, 9, 9, 0, 0, 7, 7, 8, 8, 4, 4, 6, 6, 11, 2, 2, 10, 1, 1, 3, 3, 13, 5} & 0.104 \\
15 & \small{12, 9, 9, 14, 8, 7, 7, 3, 3, 4, 4, 6, 6, 11, 2, 2, 10, 0, 0, 1, 1, 13, 5} & 0.085 \\
16 & \small{12, 9, 9, 14, 8, 3, 3, 1, 1, 4, 4, 6, 6, 11, 2, 2, 10, 15, 7, 0, 0, 13, 5} & 0.069 \\
17 & \small{12, 9, 9, 14, 8, 1, 1, 0, 0, 4, 4, 6, 6, 11, 2, 2, 10, 15, 16, 7, 3, 13, 5} & 0.055 \\
18 & \small{12, 4, 4, 14, 17, 0, 0, 8, 9, 1, 1, 6, 6, 11, 2, 2, 10, 15, 16, 7, 3, 13, 5} & 0.043 \\
19 & \small{12, 4, 4, 14, 17, 18, 13, 8, 9, 1, 1, 2, 2, 11, 0, 0, 10, 15, 16, 7, 3, 6, 5} & 0.032 \\
20 & \small{12, 1, 1, 14, 17, 18, 13, 8, 19, 0, 0, 2, 2, 11, 9, 10, 4, 15, 16, 7, 3, 6, 5} & 0.022 \\
21 & \small{12, 1, 1, 14, 17, 18, 13, 8, 19, 20, 11, 0, 0, 5, 9, 10, 4, 15, 16, 7, 3, 6, 2} & 0.014 \\
22 & \small{12, 0, 0, 14, 17, 18, 13, 8, 19, 20, 11, 21, 16, 5, 9, 10, 4, 15, 7, 3, 1, 6, 2} & 0.007 \\
\bottomrule
\end{tabular}%
}
\caption{Layer groups for every $G$ up to $L-1$ for Pythia-410M}
\label{tab:pythia-410m-groups}
\end{table}

\begin{table}[h]
\centering
\resizebox{.4\textwidth}{!}{%
\begin{tabular}{lcl}
\toprule
\textbf{G} & \textbf{Groups} & \textbf{{AMAD}} \\
\midrule
1  & \small{0, 0, 0, 0, 0, 0, 0, 0, 0, 0, 0, 0, 0, 0, 0}                & 0.459 \\
2  & \small{0, 0, 0, 0, 0, 0, 0, 0, 0, 1, 1, 1, 1, 1, 1}                & 0.364 \\
3  & \small{1, 1, 1, 1, 1, 2, 2, 2, 2, 0, 0, 0, 0, 0, 0}                & 0.309 \\
4  & \small{0, 0, 0, 0, 0, 2, 2, 2, 2, 1, 1, 1, 1, 3, 3}                & 0.250 \\
5  & \small{4, 4, 1, 1, 1, 2, 2, 2, 2, 0, 0, 0, 0, 3, 3}                & 0.225 \\
6  & \small{4, 4, 1, 1, 1, 0, 0, 0, 0, 2, 2, 5, 5, 3, 3}                & 0.191 \\
7  & \small{1, 1, 0, 0, 0, 4, 4, 2, 2, 6, 6, 5, 5, 3, 3}                & 0.174 \\
8  & \small{0, 0, 7, 3, 3, 4, 4, 2, 2, 6, 6, 5, 5, 1, 1}                & 0.139 \\
9  & \small{8, 4, 7, 3, 3, 1, 1, 2, 2, 6, 6, 5, 5, 0, 0}                & 0.100 \\
10 & \small{8, 9, 7, 1, 1, 0, 0, 2, 2, 6, 6, 5, 5, 4, 3}                & 0.075 \\
11 & \small{8, 9, 7, 0, 0, 10, 3, 2, 2, 6, 6, 5, 5, 4, 1}               & 0.053 \\
12 & \small{8, 9, 7, 11, 6, 10, 3, 0, 0, 2, 2, 5, 5, 4, 1}               & 0.036 \\
13 & \small{8, 9, 7, 11, 6, 10, 3, 12, 5, 0, 0, 2, 2, 4, 1}              & 0.022 \\
14 & \small{8, 9, 7, 11, 13, 10, 3, 12, 5, 6, 2, 0, 0, 4, 1}              & 0.010 \\
\bottomrule
\end{tabular}%
}
\caption{Layer groups for every $G$ up to $L-1$ for Pythia-1b}
\label{tab:pythia-1b-groups}
\end{table}

\newpage
\clearpage
\section{Auto Interpretability}
\label{app:auto-interp}
To evaluate the interpretability of features of baseline and Group \acp{sae}, we adopt automated pipeline from \cite{paulo2024auto-interp}, focusing on \textit{detection} and \textit{fuzzing} scores. First, an \textit{explainer} language model (LM) generates natural language explanations of the SAE latent representations. Then, a separate \textit{scorer} LM evaluates these explanations.

Then, detection scoring assesses whether a language model can identify entire sequences that activate a specific latent, given its interpretation. This method evaluates the model’s ability to distinguish between activating and non-activating contexts, offering insights into the precision and recall of the interpretation. Fuzzing scoring, on the other hand, operates at the token level, prompting the model to pinpoint specific tokens within sequences that trigger latent activations. This approach closely mirrors simulation scoring and is particularly effective in evaluating the model’s token-level understanding of latent activations.

In our experiments, we use \texttt{gemini-2.0-flash-001} as the base model for both the explainer and the scorer. For each SAE, we randomly select 64 features and cache their latent activations across 10M tokens from the Pile~\cite{pile}. To generate annotations, we present the explainer with 20 distinct examples per feature—10 that activate the latent and 10 randomly sampled—each comprising 32 tokens.

\begin{figure*}[ht!]
    \centering
    \includegraphics[width=0.8\linewidth]{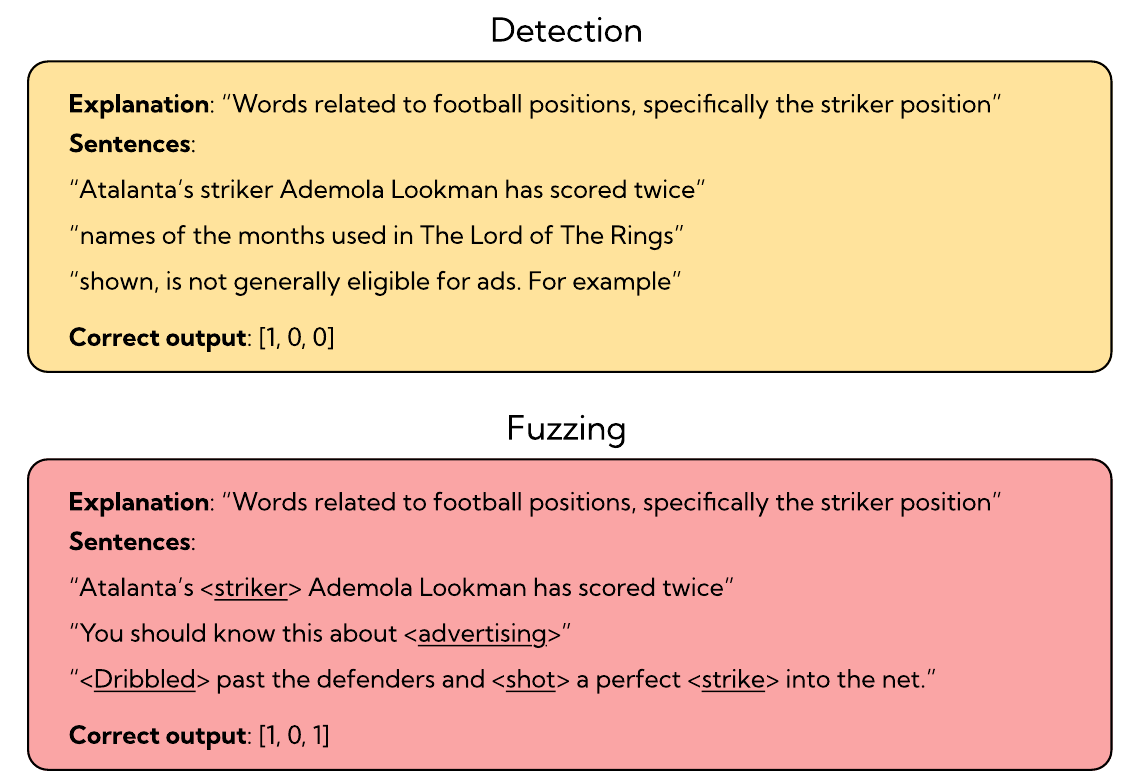}
    \caption{Examples of each of the auto-interpretability techniques: Detection and Fuzzing. In detection, the objective is to find the sentences in which the feature is active. In fuzzing, the objective is to spot the highlighted tokens referring to the target feature.}
    \label{fig:auto-interp-examples}
\end{figure*}

\newpage
\clearpage
\section{Feature Similarity Analysis}
\label{app:f_sim}
Following \citet{sharkey2023temperature}, we adopt Mean Maximum Cosine Similarity (MMCS) to assess the extent to which Baseline and Group SAEs learn similar feature directions. For any two SAEs, $\text{SAE}_i$ and $\text{SAE}_j$, we compute the MMCS between their decoder matrices $\mathbf{W}^i_{d}, \mathbf{W}^j_{d}\in\mathbb{R}^{n\times m}$ as these matrices encode the directions of the learned features:
\begin{equation}
\text{MMCS}(\mathbf{W}^i_{d},\mathbf{W}^j_{d})=\frac{1}{m} \sum_{k=1}^{m} \max_{l\in\{1,...,m\}} \cos(\widetilde{\mathbf{w}}^i_k, \widetilde{\mathbf{w}}^j_l)
\label{eq:mmcs}
\end{equation}
where $\widetilde{\mathbf{w}}^i_k$ and $\widetilde{\mathbf{w}}^j_l$ are the $k$-th and $l$-th columns of the normalized decoder matrices $\widetilde{\mathbf{W}}^i_{d}$ and $\widetilde{\mathbf{W}}^j_{d}$, respectively.
The directionality of the maximum operation is important for interpretation: we first find, for each feature in $\text{SAE}_i$, the most similar feature in $\text{SAE}_j$ (by cosine similarity), and then average these maximum similarities across all features of $\text{SAE}_i$. In our analysis, we specifically compute $\text{MMCS}(\mathbf W^\text{Baseline}_d, \mathbf W^\text{Group}_d)$, meaning that the resulting value represents the average highest similarity that each Baseline SAE feature has with any feature in the Group SAE.

\begin{figure*}[ht!]
    \centering
    \includegraphics[width=\textwidth]{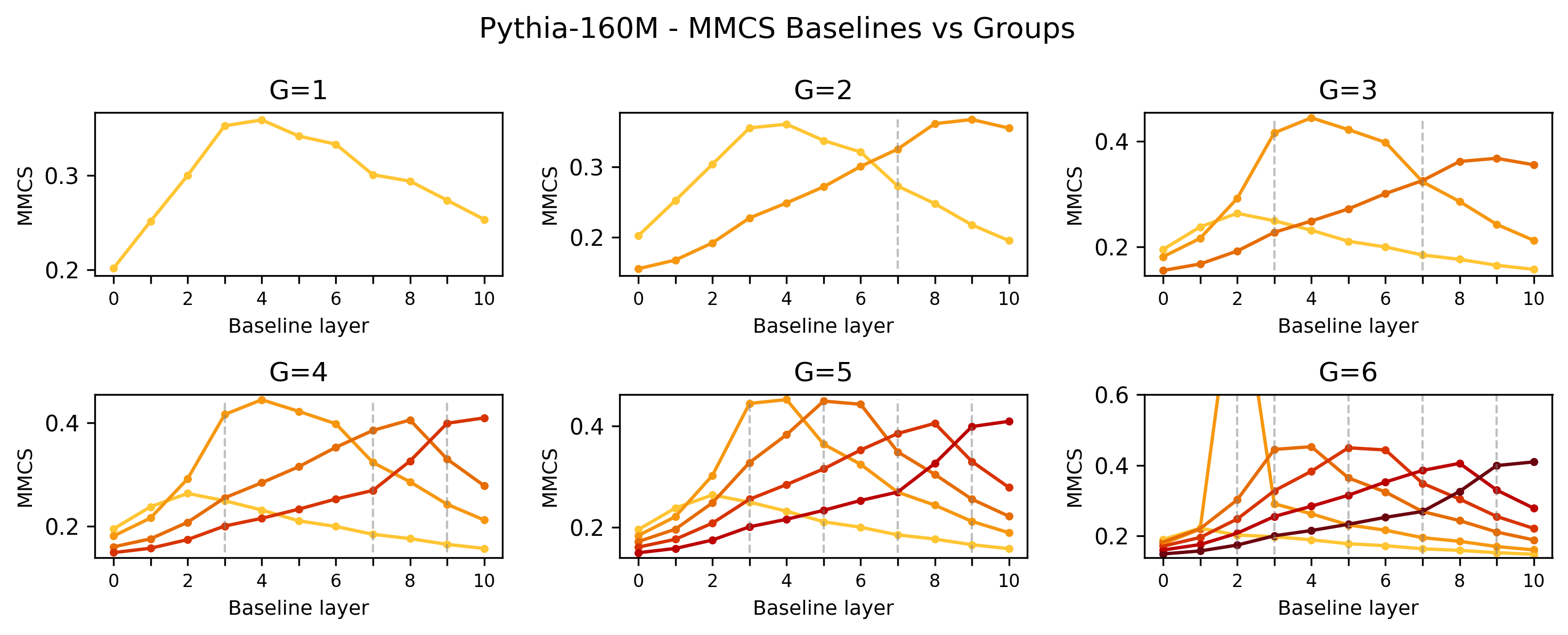}
    \caption{{Mean Maximum Cosine Similarity (MMCS) between all the learned features of baseline and group SAEs for each group $G \in \{1,\dots,\widehat{G}\}$ of Pythia-160M. Colors represent the different Group SAEs of a given partition.}
}    \label{fig:mmcs_pythia-160m}
\end{figure*}

\begin{figure*}[ht!]
    \centering
    \includegraphics[width=\textwidth]{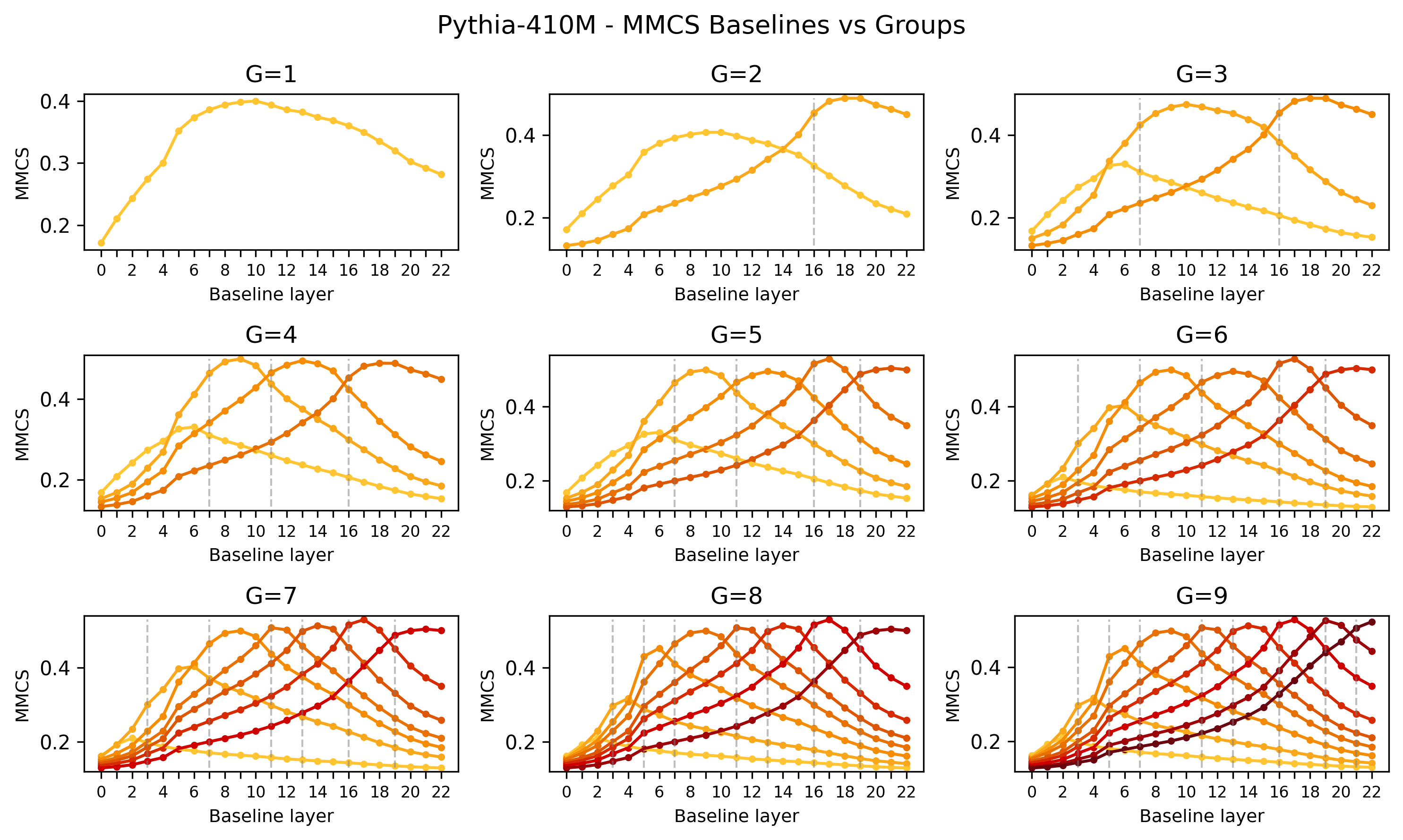}
    \caption{{Mean Maximum Cosine Similarity (MMCS) between all the learned features of baseline and Group SAEs for each group $G \in \{1,\dots,\widehat{G}\}$ of Pythia-410M. Colors represent the different Group SAEs of a given partition.}
}    \label{fig:mmcs_pythia-410m}
\end{figure*}

\begin{figure*}[ht!]
    \centering
    \includegraphics[width=\textwidth]{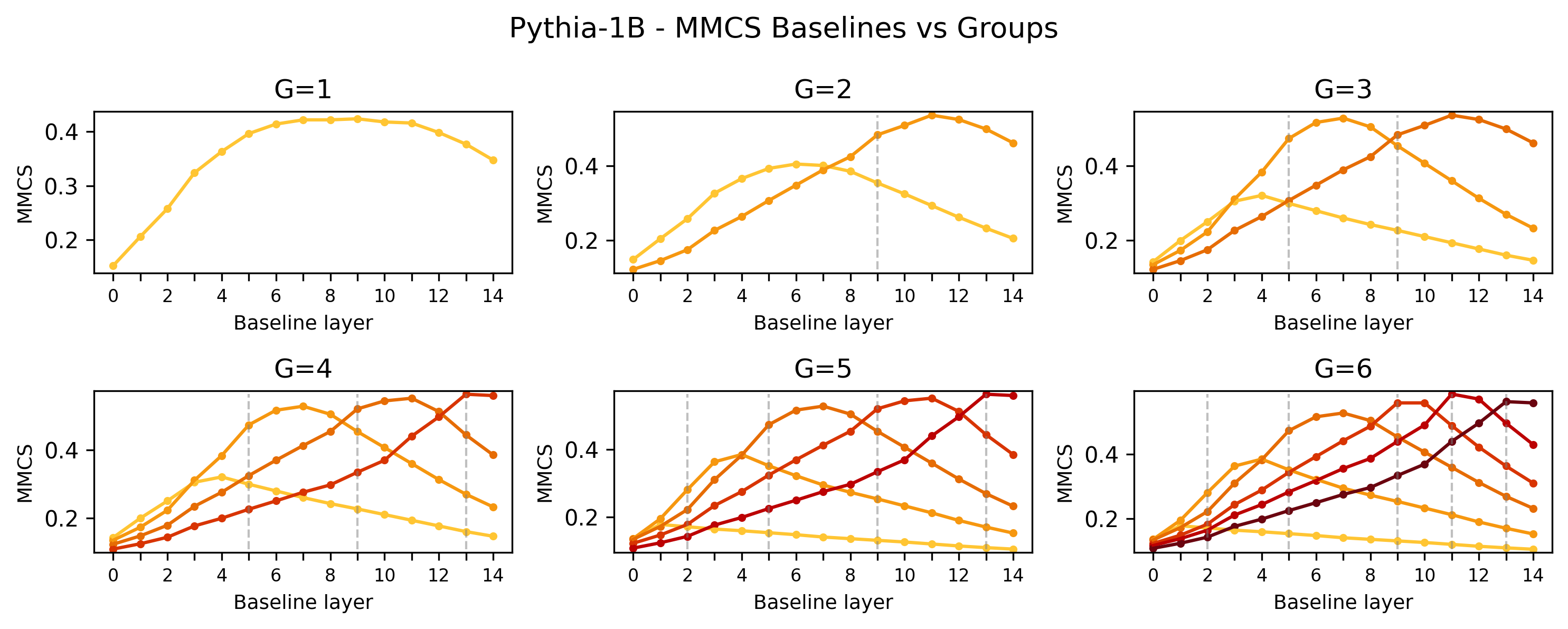}
    \caption{{Mean Maximum Cosine Similarity (MMCS) between all the learned features of baseline and Group SAEs for each group $G \in \{1,\dots,\widehat{G}\}$ of Pythia-1B. Colors represent the different Group SAEs of a given partition.}
}
\label{fig:mmcs_pythia-1b}
\end{figure*}

\newpage
\clearpage
\section{Feature Distribution Analysis}
\label{app:f_dist}
Following \cite{lawson2024mlsae}, we perform a study to understand how features distribute across layers of a given group. Previous work from \cite{lindsey2024crosscoders} showed that activations of a given feature usually peak at a specific layer. To measure this phenomenon, for each Group SAE of a given partition in $G$ groups, we sample 1 million tokens from the test set and compute feature distributions across the layers of its group.

\begin{figure*}[hb!]
    \centering
    \includegraphics[width=\linewidth]{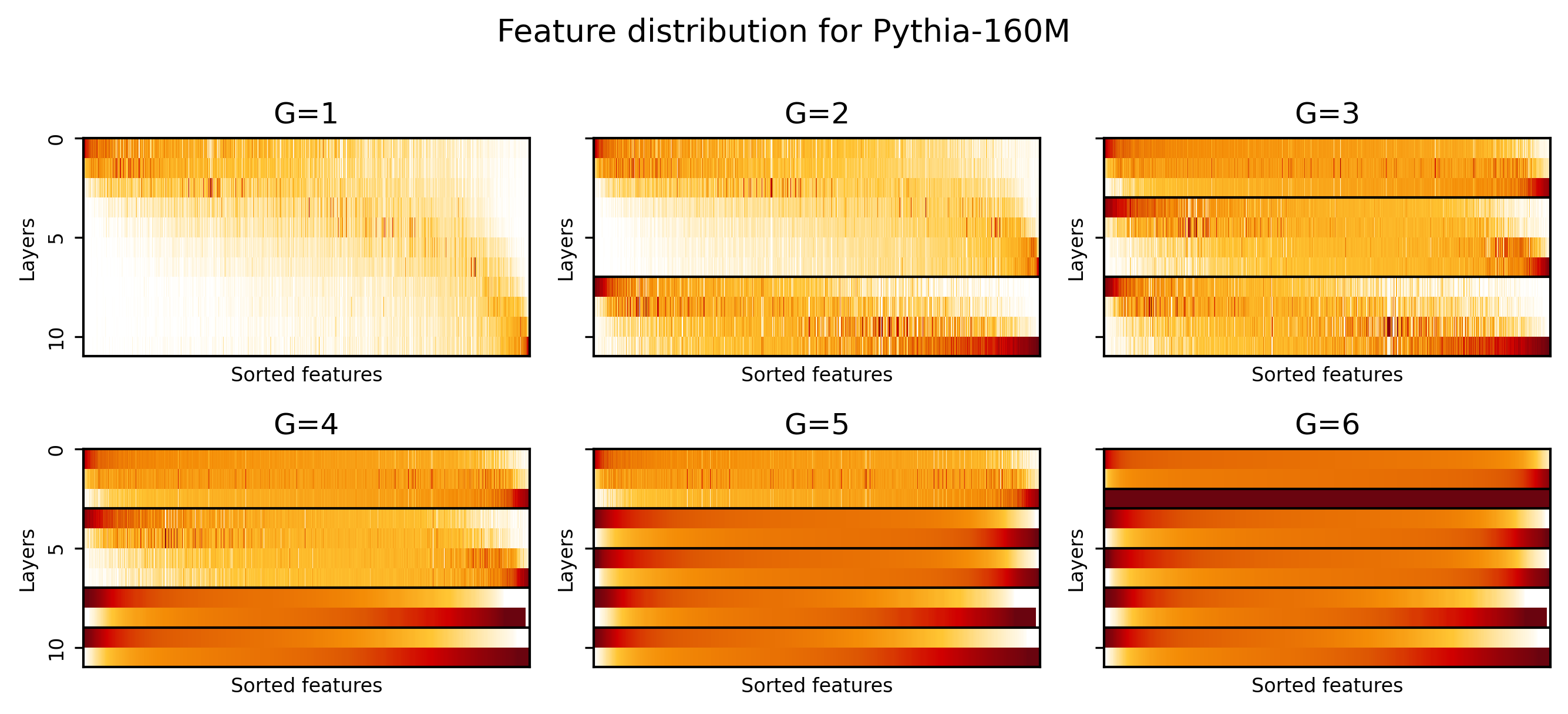}
    \caption{{Pythia-160M feature activations distribution for every group $G \in \{1,..., \widehat{G}\}$ over 1 million tokens from the test set. Darker regions indicate higher feature activation density.}}
    \label{fig:feat_spread_pythia-160m}
\end{figure*}

\begin{figure*}[ht!]
    \centering
    \includegraphics[width=\textwidth]{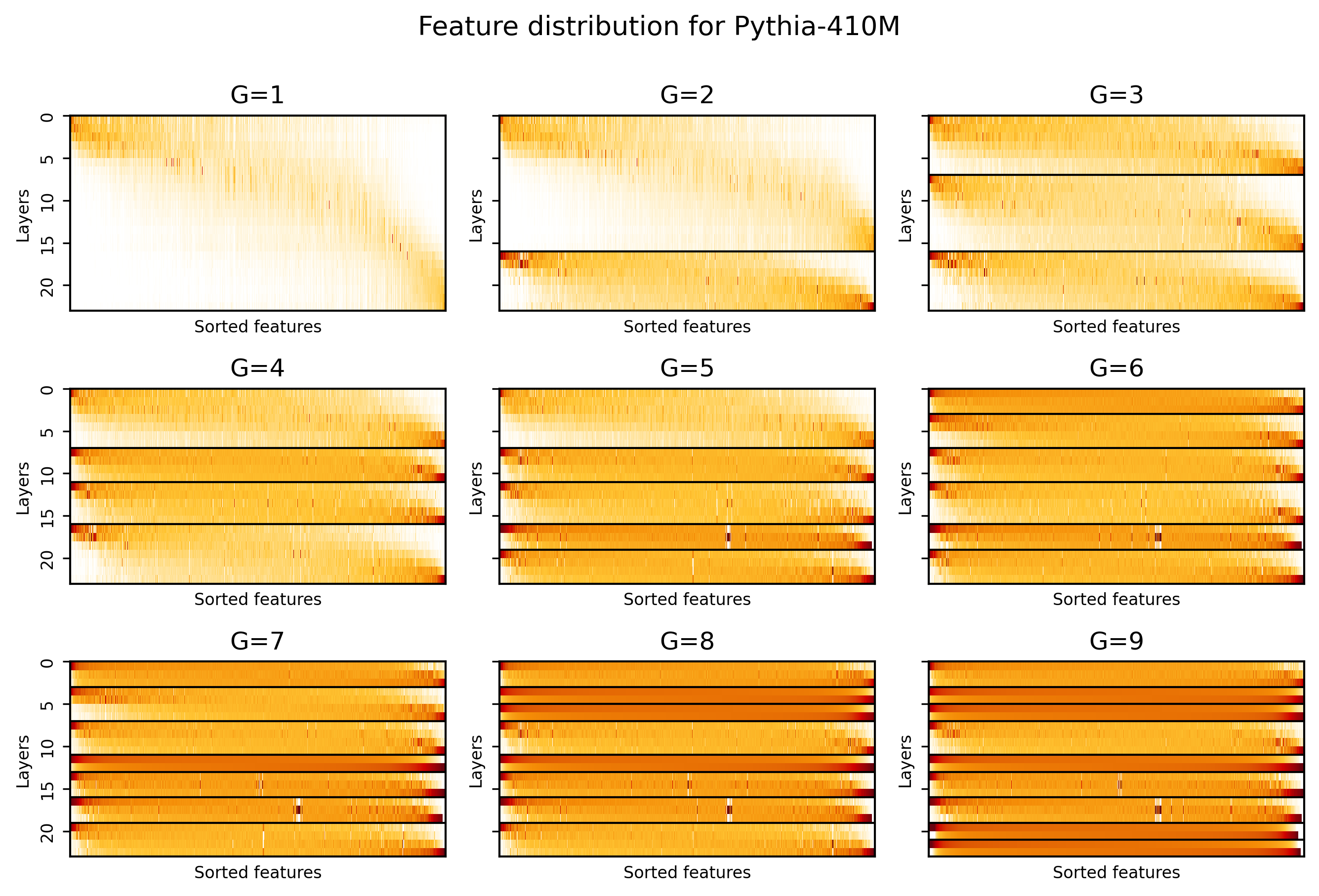}
    \caption{{Pythia-410M feature activations distribution for every group $G \in \{1,..., \widehat{G}\}$ over 1 million tokens from the test set. Darker regions indicate higher feature activation density.}}
    \label{fig:feat_spread_pythia-410m}
\end{figure*}

\begin{figure*}[ht!]
    \centering
    \includegraphics[width=\textwidth]{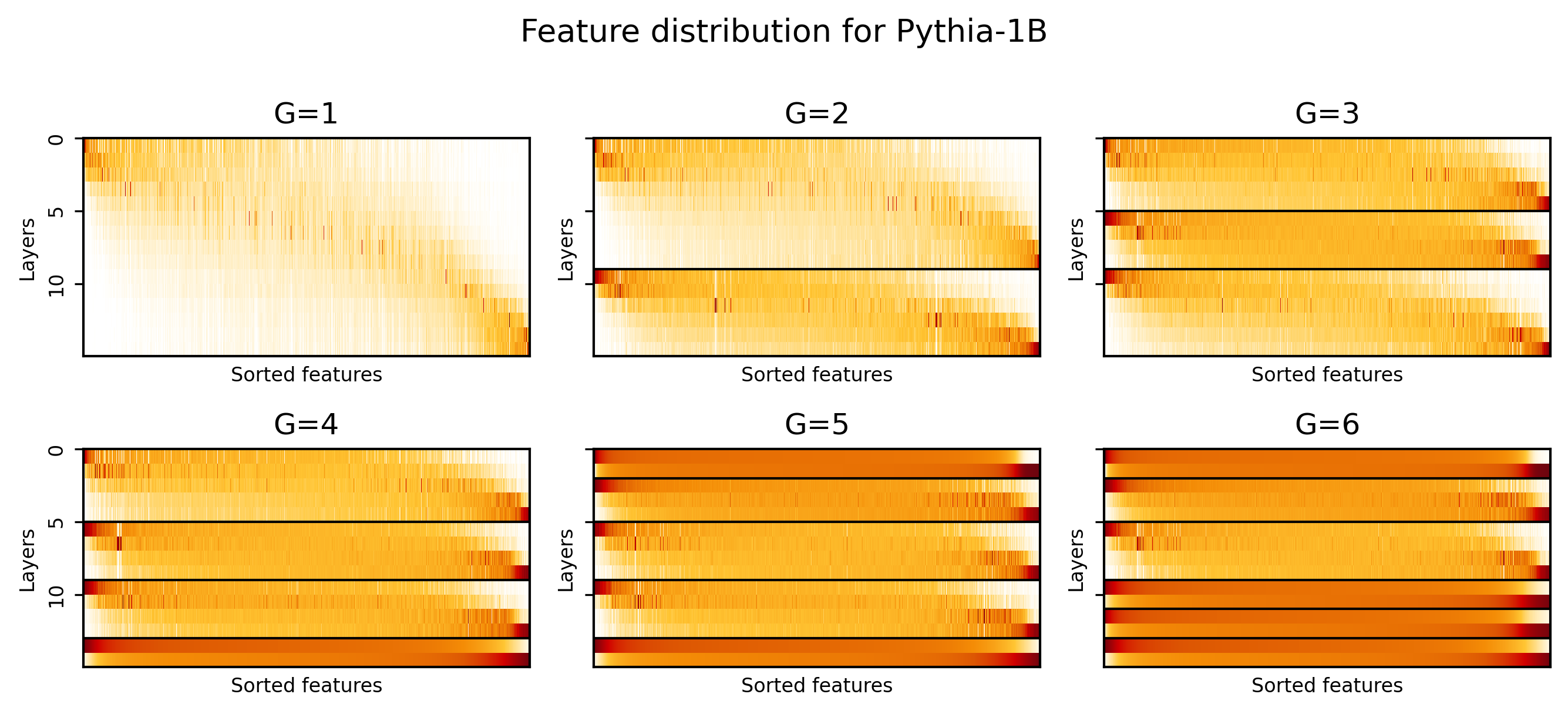}
    \caption{{Pythia-1b feature activations distribution for every group $G \in \{1,..., \widehat{G}\}$ over 1 million tokens from the test set. Darker regions indicate higher feature activation density.}}
    \label{fig:feat_spread_pythia-1b}
\end{figure*}

Heatmaps in Figure \ref{fig:feat_spread_pythia-160m}, \ref{fig:feat_spread_pythia-410m}, and \ref{fig:feat_spread_pythia-1b} show distributions of features activations for all the models and Group SAEs of partitions from 1 to $\widehat{G}$. In the images, we sort the features by the average layer they activate the most. Darker regions indicate higher feature activation density. Looking at the charts several considerations can be drawn:
\begin{itemize}
    \item Features activating for the first and last layers of a given group tend to be more specific for that layers (i.e. their activation frequencies peak at those layers).
    \item Features at early layers of a model are more spread across their respective group.
    \item Bigger models tend to have features more spread across the layers of a given group with respect to smaller models.
\end{itemize}
In summary, while feature distributions tend to peak at a specific layer (with this being more evident in smaller models and later layers), they also spread across close ones. This result agrees with findings from \cite{lindsey2024crosscoders} while still leaving the potential for Group SAEs to make SAE training more efficient.

\input{acronym}
\end{document}

%% file: acronym.tex
\begin{acronym}
\acro{ai}[AI]{Artificial Intelligence}
\acroplural{ai}[AI systems]{Artificial Intelligences}

\acro{dl}[DL]{Deep Learning}
\acroplural{dl}[DL methods]{Deep Learning methods}

\acro{sota}[SoTA]{State-of-the-Art}
\acroplural{sota}[SoTA techniques]{State-of-the-Art techniques}

\acro{nn}[NN]{Neural Network}
\acroplural{nn}[NNs]{Neural Networks}

\acro{llm}[LLM]{Large Language Model}
\acroplural{llm}[LLMs]{Large Language Models}

\acro{mi}[MI]{Mechanistic Interpretability}
\acroplural{mi}[MI studies]{Mechanistic Interpretability studies}

\acro{sae}[SAE]{Sparse Autoencoder}
\acroplural{sae}[SAEs]{Sparse Autoencoders}

\acro{nlp}[NLP]{Natural Language Processing}
\acroplural{nlp}[NLP tasks]{Natural Language Processing tasks}

\acro{cels}[CE Loss Score]{Cross-Entropy Loss Score}
\acroplural{cels}[CE Loss Scores]{Cross-Entropy Loss Scores}

\acrodef{delcels}[$\Delta$CES]{Delta Cross-Entropy Loss Score}
\acroplural{delcels}[$\Delta$CES values]{Delta Cross-Entropy Loss Scores}

\acro{lrh}[LRH]{Linear Representation Hypothesis}
\acro{sh}[SH]{Superposition Hypothesis}
\end{acronym}